\documentclass[11pt,a4paper]{article}
\usepackage[hyperref]{emnlp2020}
\usepackage{times}
\usepackage{latexsym}

\usepackage{amsmath,amsfonts,amssymb}
\usepackage{graphicx}
\usepackage{textcomp}
\usepackage{gensymb}
\usepackage{enumitem}
\usepackage{multirow}
\usepackage{caption}
\usepackage{subcaption}

\usepackage{microtype}

\aclfinalcopy 

\newcommand{\dataname}{\textsc{ArraMon}}

\title{\textsc{ArraMon}: A Joint Navigation-Assembly Instruction Interpretation Task in Dynamic Environments}

\author{Hyounghun Kim \;\;\;
Abhay Zala \;\;\;
Graham Burri \;\;\;
Hao Tan \;\;\;
Mohit Bansal \\
UNC Chapel Hill \\
\{hyounghk, aszala, ghburri, airsplay, mbansal\}@cs.unc.edu}

\date{}

\graphicspath{{figures/}}

\begin{document}
\maketitle
\begin{abstract}

For embodied agents, navigation is an important ability but not an isolated goal. Agents are also expected to perform specific tasks after reaching the target location, such as picking up objects and assembling them into a particular arrangement. 
We combine Vision-and-Language Navigation, assembling of collected objects, and object referring expression comprehension, to create a novel joint navigation-and-assembly task, named \dataname{}. During this task, the agent (similar to a Poké\textsc{Mon} GO player) is asked to find and collect different target objects one-by-one by navigating based on natural language instructions in a complex, realistic outdoor environment, but then also \textsc{Arra}nge the collected objects part-by-part in an egocentric grid-layout environment.
To support this task, we implement a 3D dynamic environment simulator and collect a dataset (in English; and also extended to Hindi) with human-written navigation and assembling instructions, and the corresponding ground truth trajectories.
We also filter the collected instructions via a verification stage, leading to a total of 7.7K task instances (30.8K instructions and paths). We present results for several baseline models (integrated and biased) and metrics (nDTW, CTC, rPOD, and PTC), and the large model-human performance gap demonstrates that our task is challenging and presents a wide scope for future work.\footnote{Our dataset, simulator, and code are publicly available at: \url{https://arramonunc.github.io}}
\end{abstract}

\section{Introduction} 

\begin{figure}[t]
    \centering
    \includegraphics[width=0.98\columnwidth]{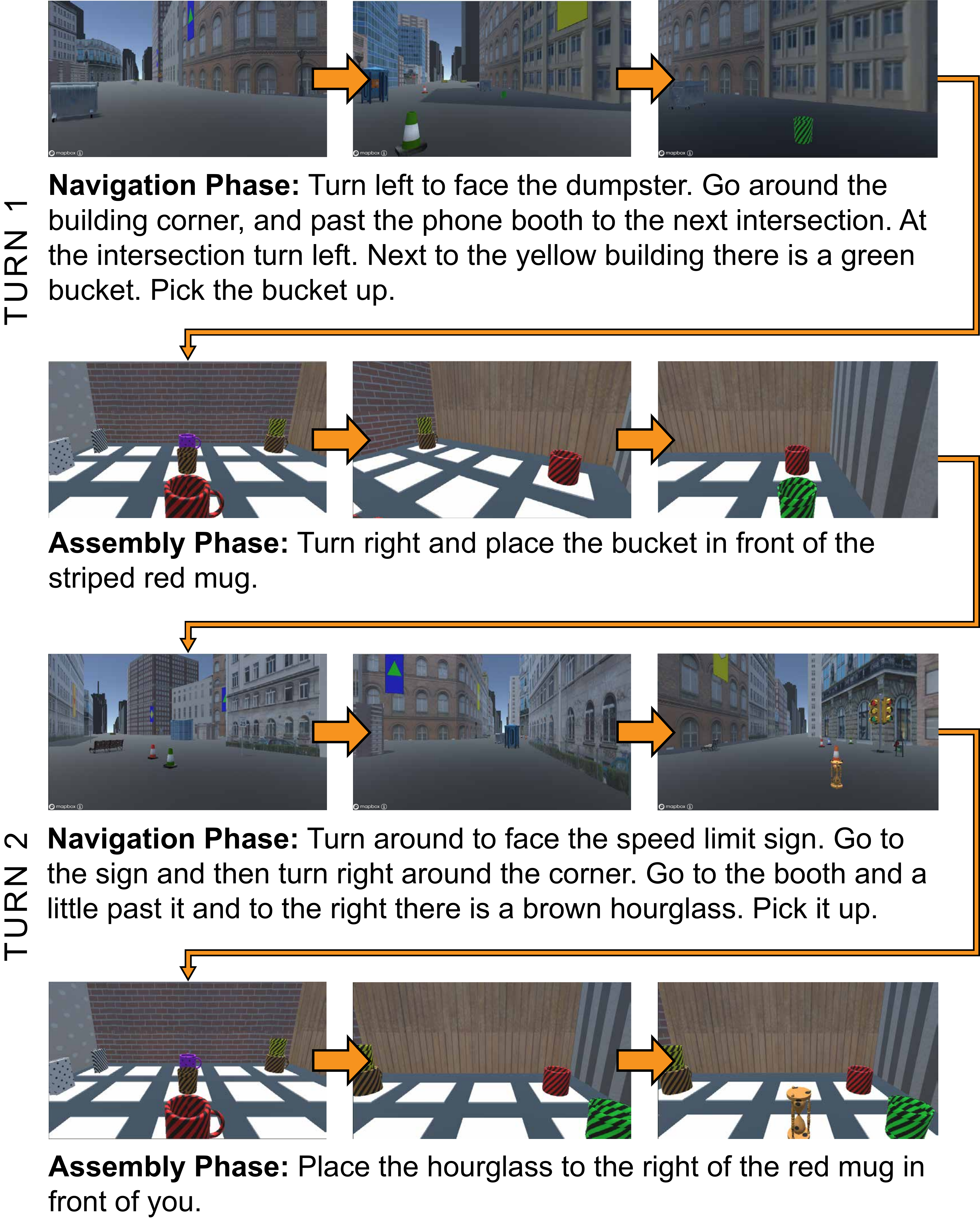}
    \vspace{-7pt}
    \caption{Navigation and assembly phases (2 turns), via NL (English) instructions in a dynamic 3D environment. In the navigation phase, agents are asked to find and collect a target object. In the assembly phase, agents have to egocentrically place the collected object at a relative location (navigation turn 2 starts where turn 1 ends; we only show 3 snapshots here for space reasons, but the full simulator and its image set will be made available).}
    \label{fig:two_tasks}
\vspace{-11pt}
\end{figure}

Navigation guided via flexible natural language (NL) instructions is a crucial capability for robotic and embodied agents. Such systems should be capable of interpreting human instructions to correctly navigate realistic complex environments and reach destinations by understanding the environment, and associating referring expressions in the instructions with the corresponding visual cues in the environment.
Many research efforts have focused on this important vision-and-language navigation task~\cite{macmahon2006walk, mooney2008learning, chen2011learning,tellex2011understanding, mei2016listen,hermann2017grounded, brahmbhatt2017deepnav, mirowski2018learning,  anderson2018room2room, misra2018mapping, blukis2018mapping, embodiedqa, cirik2018following, de2018talk, blukis2020learning, thomason:corl19, nguyen2019vnla, nguyen2019hanna,  Chen19:touchdown, sotp2019acl, ALFRED20, qi2019rerere, hermann2019learning, berg2020grounding, Zhu2020BabyWalkGF}. However, in real-world applications, navigation alone is rarely the exclusive goal. In most cases, agents will navigate to perform another task at their destination, and also repeat subtasks, e.g., a warehouse robot may be asked to pick up several objects from different locations and then assemble the objects into a desired arrangement. When these additional tasks are interweaved with navigation, the degree of complexity increases exponentially due to cascading errors. Relatively few studies have focused on this idea of combining navigation with other tasks. Touchdown~\cite{Chen19:touchdown} combines navigation and object referring expression resolution, REVERIE~\cite{qi2019rerere} performs remote referring expression comprehension, and ALFRED~\cite{ALFRED20} combines indoor navigation and household manipulation. However, there has been no task that integrates the navigation task in complex outdoor spaces with the assembling task (and object referring expression comprehension), requiring spatial relation understanding in an interweaved temporal way, in which the two tasks alternate for multiple turns with cascading error effects (see Figure~\ref{fig:two_tasks}).

Thus, we introduce a new task that combines the navigation, assembling, and referring expression comprehension subtasks. This new task can be explained as an intuitive combination of the navigation and collection aspects of Poké\textsc{Mon} GO\footnote{\url{https://www.pokemongo.com}} and an \textsc{Arra}nging (assembling) aspect, hence we call it `\dataname{}'. In this task, an agent needs to follow navigational NL instructions to navigate through a complex outdoor and fine-grained city environment to collect diverse target objects via referring expression comprehension and dynamic 3D visuospatial relationship understanding w.r.t. other distracter objects. Next, the agent is asked to place those objects at specific locations (relative to other objects) in a grid environment based on an assembling NL instruction. These two phases are performed repeatedly in an interweaved manner to create an overall configuration of the set of collected objects. For enabling the \dataname{} task, we also implement a simulator built in the Unity game engine\footnote{\url{https://www.unity.com}} to collect the dataset (see Appendix~\ref{app:interface} for the simulator interface). This simulator features a 3D synthetic city environment based on real-world street layouts with realistic buildings and textures (backed by Mapbox\footnote{\url{https://www.mapbox.com}}) and a dynamic grid floor assembly room (Figure \ref{fig:two_tasks}), both from an egocentric view (the full simulator and its image set will be made available). We take 7 disjoint sub-sections from the city map and collect instructions from workers within each section. Workers had to write instructions based on ground truth trajectories (represented as path lines in navigation, location highlighting during assembly). We placed diverse background objects as well as target objects so that the rich collected instructions require agents to utilize strong linguistic understanding. The instructions were next executed by a new set of annotators in a second verification stage and were filtered based on low match w.r.t. the original ground truth trajectory, and the accuracy of assembly placement. Overall, this resulted in a dataset of 7,692 task instances with multiple phases and turns (a total of 30,768 instructions and paths).\footnote{Our dataset size is comparable to other similar tasks (e.g., R2R, Touchdown, ALFRED, CVDN, REVERIE; we are also planning to further increase the size and add other languages.} We have since extended our dataset by also collecting the corresponding Hindi instructions.

To evaluate performance in our \dataname{} task, we employ both the existing metric of nDTW (Normalized Dynamic Time Warping) \cite{ilharco2019sdtw} and our newly-designed metrics: CTC-k (Collected Target Correctness), rPOD (Reciprocal Placed Object Distance), and PTC (Placed Target Correctness). In the navigation phase, nDTW measures how similar generated paths are to the ground truth paths, while CTC-k computes how closely agents reach the targets. In the assembly phase, rPOD calculates the reciprocal distance between target and agents' placement locations, and PTC counts the correspondence between those locations. Due to the interweaving property of our task with multiple navigation and assembling phases and turns, performance in the previous turn and phase cascadingly affects the metric scoring of the next turn and phase (Section \ref{sec:metrics}). 

Lastly, we implement multiple baselines as good starting points and to verify our task is challenging and the dataset is unbiased. We present integrated vision-and-language, vision-only, language-only, and random-walk baselines. Our vision-and-language model shows better performance over the other baselines, which implies that our \dataname{} dataset is not skewed; moreover, there exists a very large gap between this model and the human performance, implying that our \dataname{} task is challenging and that there is substantial room for improvements by future work. We will publicly release the \dataname{} simulator, dataset, and code, along with a leaderboard to encourage further community research on this realistic and challenging joint navigation-assembly task.

\begin{figure}[t]
    \centering
    \includegraphics[width=1\columnwidth]{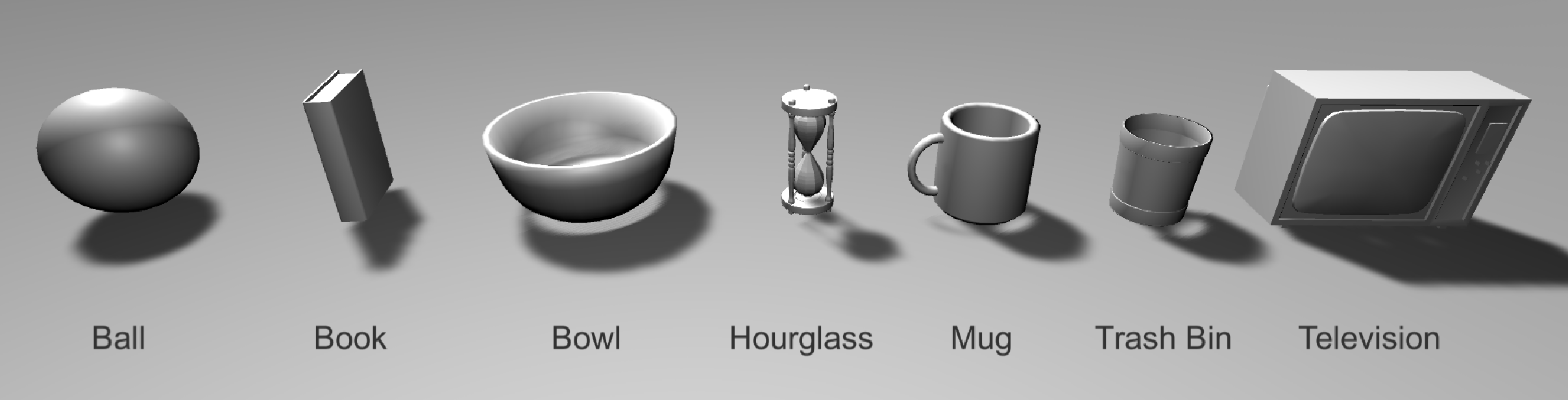}
    \caption{Illustration of the basic object types that the agent must collect, and will also appear as distracter objects during both navigation and assembly phases.}
    \label{fig:objects_graphic}
    \vspace{-10pt}
\end{figure}

\section{Related Work}
\noindent\textbf{Vision-and-Language Navigation.}
Recently, Vision-and-Language Navigation (VLN) tasks, in which agents follow NL instructions to navigate through an environment, have been actively studied in research communities \cite{macmahon2006walk, mooney2008learning, chen2011learning,tellex2011understanding, mei2016listen,hermann2017grounded, brahmbhatt2017deepnav, mirowski2018learning,  anderson2018room2room, misra2018mapping, blukis2018mapping, embodiedqa, cirik2018following, de2018talk, blukis2020learning, thomason:corl19, nguyen2019vnla, nguyen2019hanna,  Chen19:touchdown, sotp2019acl, ALFRED20, qi2019rerere, hermann2019learning, berg2020grounding, Zhu2020BabyWalkGF}. To encourage the exploration of this challenging research topic, multiple simulated environments have been introduced. Synthetic \cite{Kempka2016ViZDoom, beattie2016deepmind, kolve2017ai2, brodeur2017home, wu2018building, savva2017minos, zhu2017visual, yan2018chalet, shah2018follownet, puig2018virtualhome} as well as real-world and image-based environments \cite{brahmbhatt2017deepnav, mirowski2018learning, anderson2018room2room, xia2018gibson, cirik2018following, de2018talk, Chen19:touchdown, savva2019habitat} have been used to provide agents with diverse and complement training environments.

\noindent\textbf{Referring Expression Comprehension.}
The ability to make connections between objects or spatial regions and the natural language expressions that describe those objects or regions, has been a focus of many studies. Given that humans regularly carry out complex symbolic-spatial reasoning, there has been much effort to improve the capability of referring expression comprehension (including remote objects) in agents~\cite{kazemzadeh2014referitgame, mao2016generation, hu2016natural, yu2018mattnet, Chen19:touchdown, qi2019rerere}, but such reasoning remains challenging for current models. Our \dataname{} task integrates substantial usage of referring expression comprehension as a requirement, as it is necessary to the successful completion of both the navigation and assembly phases.

\noindent\textbf{Assembling Task.}
Object manipulation and configuration is another subject that has been studied along with language and vision grounding \cite{bisk2016towards, wang2016learning, li2016spatial, bisk2018learning}. However, most studies focus on addressing the problem in relatively simple environments from a third-person view. Our \dataname{} task, on the other hand, provides a challenging dynamic, multi-step egocentric viewpoint within a more realistic and interactive 3D, depth-based environment. Moreover, the spatial relationships in \dataname{} dynamically change every time the agent moves, making `spatial-action' reasoning more challenging. We believe that an egocentric viewpoint is a key part of how humans perform spatial reasoning, and that such an approach is therefore vital to producing high-quality models and datasets.

These three directions of research are typically pursued independently (esp. navigation and assembling), and there have been only a few recent efforts to combine the traditional navigation task with other tasks.
Touchdown~\cite{Chen19:touchdown} combines navigation and object referring expression resolution, REVERIE~\cite{qi2019rerere} performs remote referring expression comprehension, while ALFRED~\cite{ALFRED20} combines indoor navigation and household manipulation. Our new complementary task merges navigation in a complex outdoor space with object referring expression comprehension and assembling tasks that require spatial relation understanding in an interweaved temporal style, in which the two tasks alternate for multiple turns leading to cascading error effects. This will allow development of agents with more integrated, human-like abilities that are essential in real-world applications such as moving and arranging items in warehouses; collecting material and assembling structures in construction sites; finding and rearranging household objects in homes.

\section{Task}
The \dataname{} task consists of two phases: navigation and assembly. We define one turn as one navigation phase plus one assembly phase (see Figure \ref{fig:two_tasks}). Both phases are repeated twice (i.e., 2 turns), starting with the navigation phase. During the navigation phase, an agent is asked to navigate a rich outdoor city environment by following NL instructions, and then collect the target object identified in the instructions via diverse referring expressions. During the assembly phase, the agent is asked to place the collected object (from the previous navigation phase) at a target location on a grid layout, using a different NL instruction via relative spatial referring expressions. Target objects and distracter objects are selected from one of seven objects shown in Figure \ref{fig:objects_graphic} and then are given one of two different patterns and one of seven different colors (see Figure~\ref{fig:colortexture_graphic} in Appendices). In both phases, the agent can take 4 actions: forward, left, right, and an end pickup/place action. Forward moves the agent 1 step ahead and left/right makes agents rotate 30\degree~ in the respective direction.\footnote{In our task environment, holistically, the configuration of the set of objects dynamically changes as agents pick-up and place or stack them relative to the other objects, which is one challenging interaction between the objects.}
\subsection{Environment\label{sec:enviroment}}
\begin{figure}
    \centering
    \includegraphics[width=0.7\columnwidth]{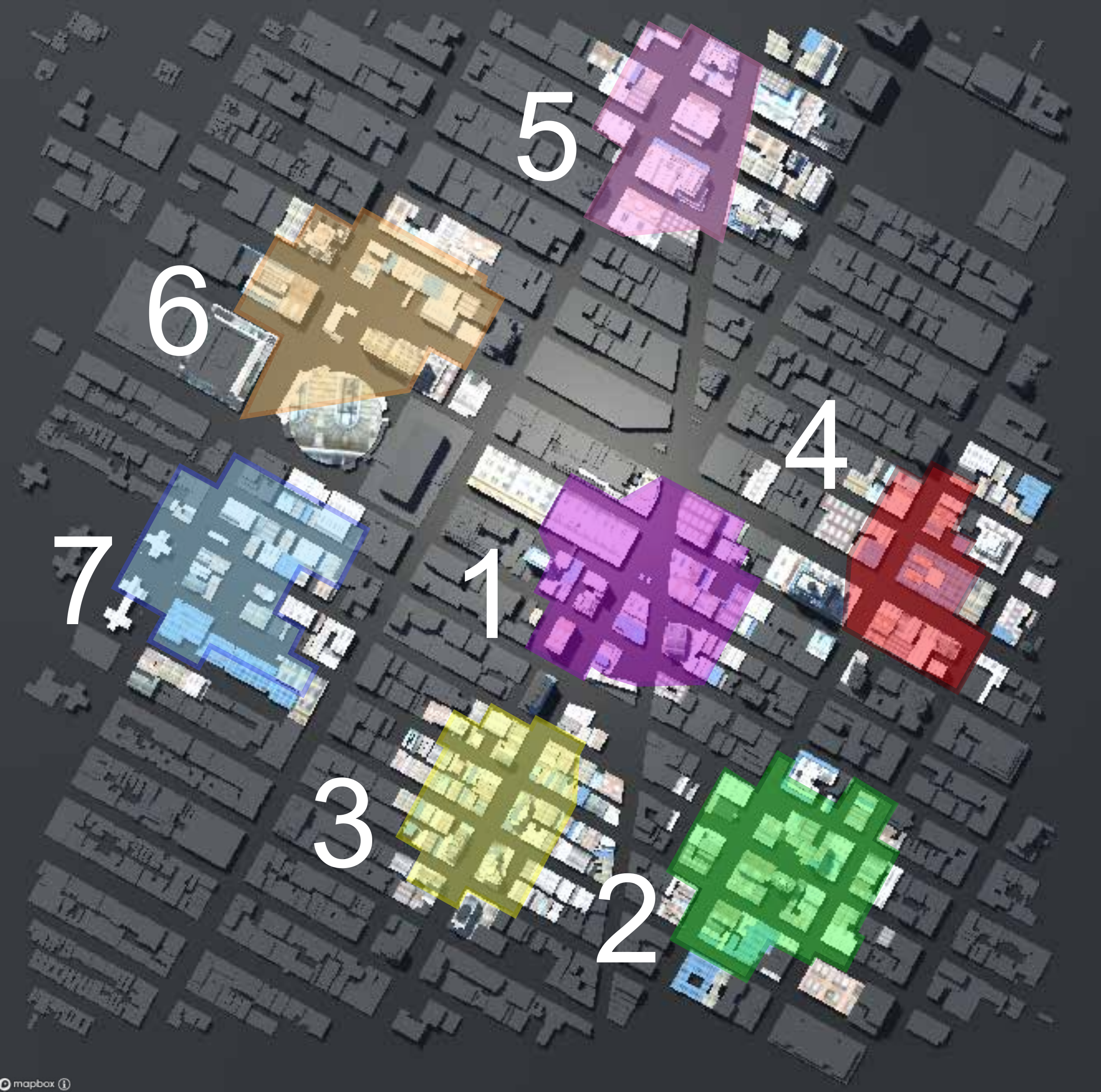}
    \caption{Illustration of the seven city sections in which data was collected.}
    \label{fig:sections}
    \vspace{-5pt}
\end{figure}

\begin{figure}[t]
    \centering
    \includegraphics[width=0.98\columnwidth]{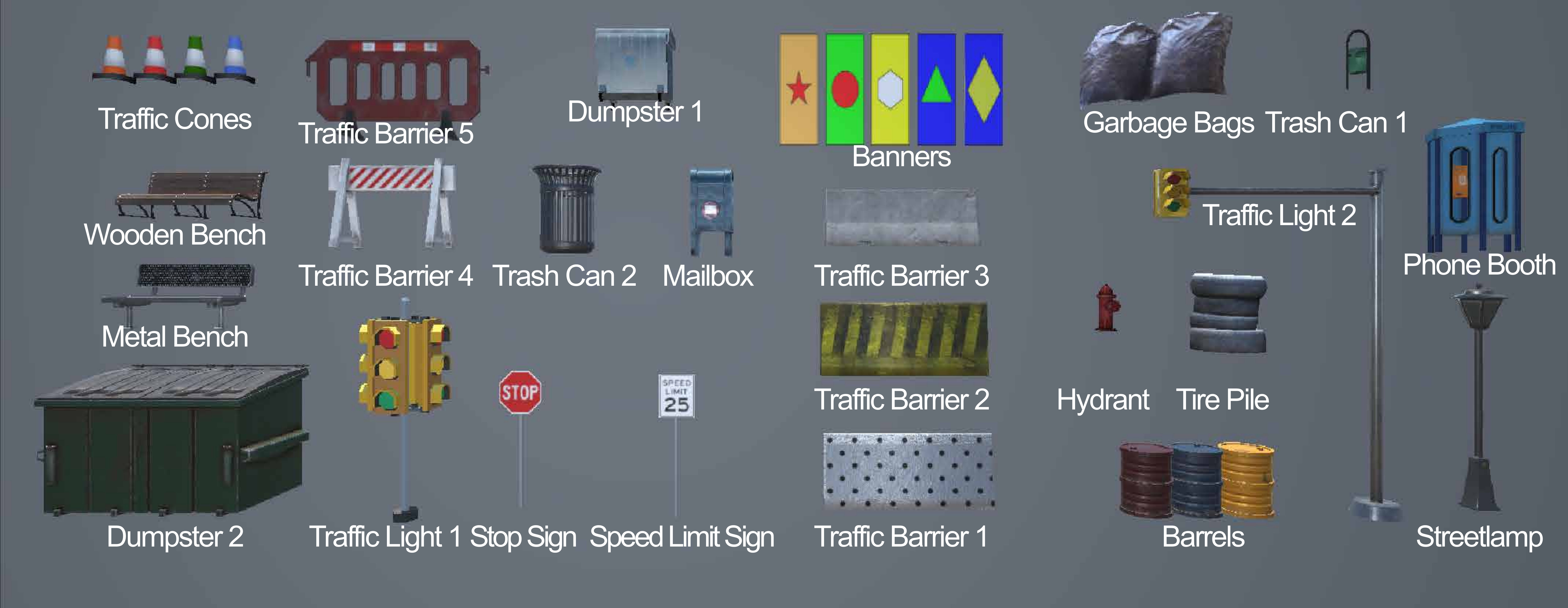}
    \vspace{-5pt}
    \caption{Illustration of the background environmental objects scattered around the city environment.}
    \label{fig:environmentobjects_graphic}
    \vspace{-10pt}
\end{figure}

\noindent\textbf{Navigation Phase.}
In this phase, agents are placed at a random spot in one of the seven disjoint subsections of the city environment (see Figure~\ref{fig:sections}), provided with an NL instruction, and asked to find the target object. The city environment is filled with background objects: buildings and various objects found on streets (see Figure \ref{fig:environmentobjects_graphic}). There are also a few distracter objects in the city that are similar to target objects (in object type, pattern, and color). During this phase, the agent's end action is `pick-up'. The pick-up action allows agents to pick up any collectible object within range (a rectangular area: 0.5 unit distance from an agent toward both their left and right hand side and 3 unit distance forward).

\noindent\textbf{Assembly Phase.}
Once the agent picks up the collectible object in the navigation phase, they enter the assembly phase. In this phase, agents are again provided with an NL instruction, but they are now asked to place the target object they collected in the previous phase at the target location identified in the instruction. When the assembly phase begins, 8 decoy basic-type objects (Figure \ref{fig:objects_graphic}) with random pattern and color, are placed for use as distractions. In this phase, agents can only move on a 4-by-5 grid layout. The grid is bordered by 4 walls, each with a different texture/pattern (wood, brick, spotted, striped) to allow for more diverse expressions in the assembly phase. Their end action is `place', which puts the collected object onto the grid one step ahead. Agents cannot place diagonally and, unlike in the navigation phase, cannot move forward diagonally.

Hence, to accomplish the overall joint navigation-assembly task, it is required for agents to have integrated abilities. During navigation they must take actions based on understanding the egocentric view and aligning the NL instructions with the dynamic visual environment to successfully find the target objects (relevant metrics: nDTW and CTC-k, see Section \ref{sec:metrics}). During assembly, from an egocentric view, they must understand 3D spatial relations among objects identified by referring expressions in order to place the target objects at the right relative location. (relevant metrics: PTC and rPOD, see Section \ref{sec:metrics}).\footnote{We assume agents backtrack their path to go back to the warehouse for assembling, after each navigation phase (since the path is known, it can be automated and there is no additional learning task involved, and so no visuals are needed). Likewise, after the assembly phase, the agent can resume at the pick-up position by re-following the previous path. One can also imagine agents are moving with a container, in which they assemble the objects as they pick them up.
}

\subsection{Metrics\label{sec:metrics}}
\noindent\textbf{Normalized Dynamic Time Warping (nDTW).}
To encourage the agent to follow the paths closely during the navigation task, we employ nDTW \cite{ilharco2019sdtw} as our task metric. nDTW measures the similarity between a ground-truth path and a predicted trajectory of an agent, thus penalizing randomly walking around to find and pick up the target object.

\noindent\textbf{Collected Target Correctness (CTC).} 
An agent that understands the given NL instructions well should find and pick up a correct target object at the end of the navigation task. Therefore, we evaluate the agent's ability with CTC, which will have a value of 1 if the agent picks up a correct object, and a value of 0 if they pick up an incorrect object or do not pick up any object.
Since collecting the correct object is a difficult task, we also implement the CTC-k metric. CTC-k measures the CTC score at distance k. If the agent is within k distance of the target object, then the value is 1, otherwise it is 0 (CTC-0 indicates the original CTC).

\noindent\textbf{Placed Target Correctness (PTC).}
In the assembly task, placing the collected object at the exact target position is most important. The PTC metric counts the correspondence between the target location and the placed location. If the placed and target locations match, then the PTC is 1, otherwise it is 0. If the collected object is not correct, then the score is also 0. 

\noindent\textbf{Reciprocal Placed Object Distance (rPOD).}
We also consider the distance between the target position and the position where the collected object is eventually placed in the assembly task \cite{bisk2018learning}. The distance squared is taken to penalize the agent more for placing the object far from the target position. Then 1 is added and the reciprocal is taken to normalize the final metric value: $\textrm{rPOD}=\frac{1}{1+D_a^2}$, where \(D_a\) is the Manhattan distance between the target and placed object positions. If the collected object is not correct, then the score is 0 (see Figure~\ref{fig:rpod} in Appendices).

Overall, our metrics reflect the interweaving property of our task. For example, if agents show poor performance in the first turn navigation phase (i.e., low nDTW and CTC-k scores), they will not obtain high scores in the continuing assembly phase (i.e., low PTC and rPOD scores), also leading to lower scores in the second turn navigation phase.

\section{\dataname{} Dataset}
Our \dataname{} navigation-assembly dataset is a collection of rich human-written NL (English) instructions. The navigation instructions explain how to navigate the large outdoor environments and describe which target objects to collect. The assembly instructions provide the desired target locations for placement relative to objects. Each instruction set in the dataset is accompanied by ground truth (GT) trajectories and placement locations. Data was collected from the online crowd-sourcing platform Amazon Mechanical Turk (AMT). 

\subsection{Data Collection \label{sec:datacollection}}
The data collection process was broken into two stages: Stage 1: Writing Instructions, and Stage 2: Following/Verifying Instructions. Within each stage, there are two phases: Navigation and Assembly (see Figure~\ref{fig:simulation_interface} in Appendices for the interface of each stage and each phase). During the first stage's navigation phase, a crowdworker is placed in the city environment as described in Section \ref{sec:enviroment} and moves along a blue navigation line (representing the GT path) that will lead them to a target object (see Appendix~\ref{app:data_collect} for the exact route generation details). While the worker travels this line, they write instructions describing their path (e.g., ``Turn to face the building with the green triangle on a blue ... Walk past the bench to the dotted brown TV and pick it up."). Workers were bound to this navigation line to ensure that they wrote instructions only based on what they could see from the GT path. Next, the worker starts the first stage's assembly phase and is placed in a small assembly room, where they must place the object they just collected in a predetermined location (indicated by a transparent black outline of the object they just collected) and write instructions on where to place the object relative to other objects from an egocentric viewpoint (e.g., ``Place the dotted brown TV in front of the striped white hourglass."). The worker is then returned to the city environment and repeats both phases once more.

A natural way of verifying the instruction sets from Stage 1 is to have new workers follow them \cite{Chen19:touchdown}. Thus, during Stage 2 Verification, a new worker is placed in the environment encountered by the Stage 1 worker and is provided with the NL instructions that were written by that Stage 1 worker. The new worker has to follow the instructions to find the target objects in the city and place them in the correct positions in the assembly environment. Each instruction set from Stage 1 is verified by three unique crowdworkers to ensure instructions are correctly verified. Next, evaluation of the Stage 2 workers performance was done through the use of the nDTW and PTC metrics. If at least one of three different Stage 2 workers scored higher than 0.2 on nDTW in both navigation turns and had a score of 1 on PTC in both assembly turns, then the corresponding Stage 1 instruction set was considered high quality and kept in the dataset, otherwise it was discarded. The remaining dataset has a high average nDTW score of 0.66 and an even higher expert score of 0.81 (see Sec.~\ref{sec:results}).\footnote{Workers were allowed to repeat both tasks, however they were prevented from encountering an identical map setting that already has instructions during Stage 1 and their own instructions during Stage 2.}

\subsection{Data Quality Control}
Instructions written by the Stage 1 workers needed to be clear and understandable. Workers were encouraged to follow certain rules and guidelines so that the resulting instruction would be of high quality and made proper use of the environment.

\begin{figure}[t]
    \centering
    \includegraphics[width=0.75\columnwidth]{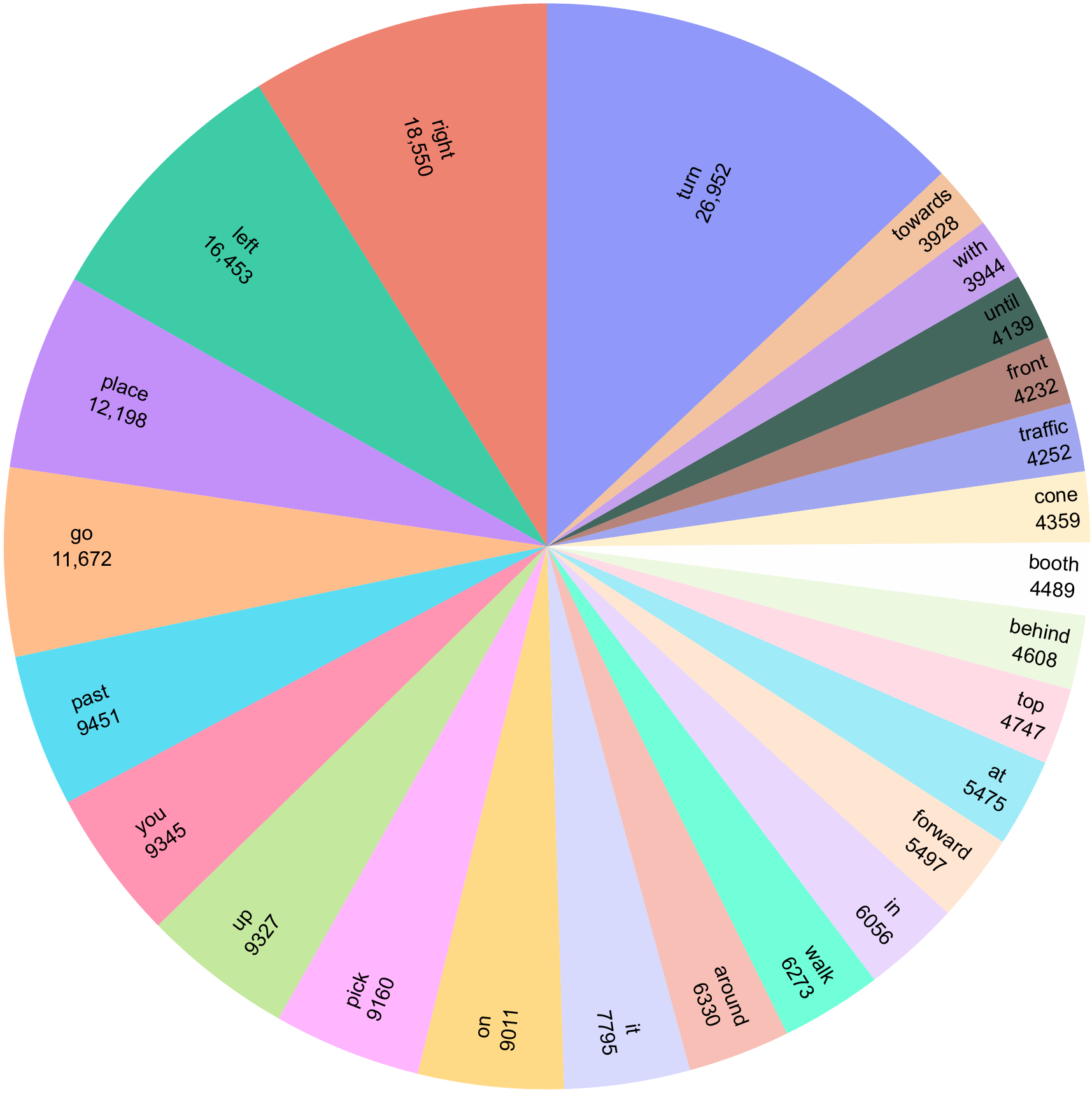}
    \vspace{-4pt}
    \caption{The frequency distribution of the 25 most common words in the dataset. Stopwords and target object words have been removed.}
    \label{fig:itempie}
    \vspace{-5pt}
\end{figure}

\noindent\textbf{Guidelines, Automated Checks, and Qualification Tests.}
Detailed guidelines were put in place to help ensure that the instructions written contained as few errors as possible. Rules were shown to workers before the start of the task and active automated checks took place as the workers wrote. These active checks helped prevent poor instructions (such as those including certain symbols) from being submitted, requiring workers to fix them before submitting. In the case the instruction quality was questionable, an email notification was sent (see Appendix~\ref{app:data_collect} for the exact guidelines and checks that were implemented, as well as details regarding the email notifications). A screening test was also required at the start of both stages to test the crowdworkers' understanding of the task. If a wrong answer was chosen, an explanation was displayed and the crowdworker was allowed to try again (see Figure~\ref{fig:quiz_stage1} and \ref{fig:quiz_stage2} in Appendices for the screening tests). To help workers place the object in the right location during Stage 2, we use a simple placement test which they pass by placing an object at the correct place during a mock assembly phase (see Appendix~\ref{app:data_collect} for details).

\noindent\textbf{Worker Qualifications.}
Workers completing the task were required to pass certain qualifications before they could begin. As the Stage 1 and 2 tasks require reading English instructions (Stage 1 also involves writing), we required workers be from native-speaking English countries. Workers were required to have at least 1000 approved tasks and a 95\% or higher approval rating. A total of 96 unique workers for Stage 1 and 242 for Stage 2, were able to successfully complete their respective tasks.

\noindent\textbf{Worker Payment and Bonus Incentives.}
We kept fair and comparable pay rates based on similar datasets \cite{Chen19:touchdown}, writing (Stage 1) had a payment (including bonuses) of \$1.00. Instruction verification (Stage 2) had a payment of \$0.20. See Appendix~\ref{app:data_collect} for details on bonus criteria, rates.

\begin{figure}[t]
    \centering
    \includegraphics[width=0.99\columnwidth\vspace{3pt}]{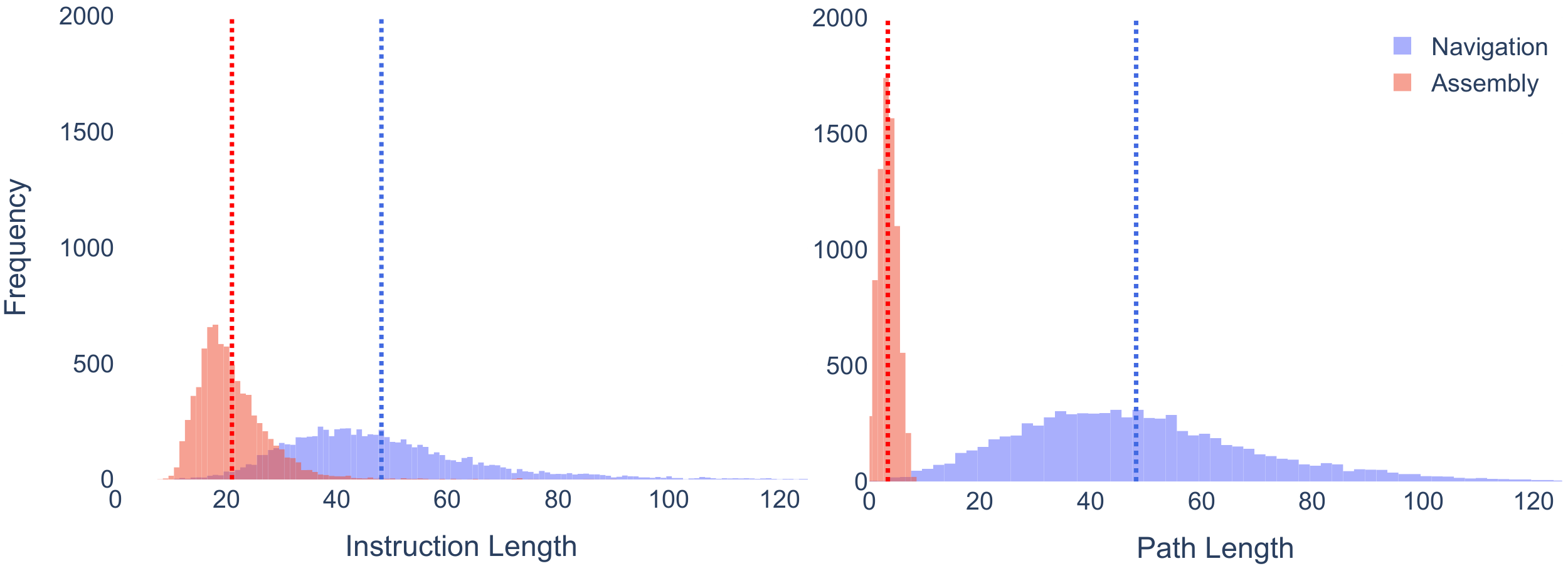}
    \vspace{-22pt}
    \caption{The frequency distributions of instruction lengths (left) and path lengths (right) in the navigation and assembly phases. Graphs cut off at length 125 since beyond that there are very few data points.}
    \label{fig:wordcounts}
    \vspace{-1pt}
\end{figure}

\section{Data Analysis}
A total of 8,546 instruction sets were collected. Each set included two pairs of navigation and assembly instructions (thus, 34,184 instructions in total). After filtering from Stage 2 results, there remained 7,692 instruction sets (30,768 instructions in total). Our dataset size is comparable to other similar tasks, e.g., Touchdown~\cite{Chen19:touchdown} contains 9.3K examples (9.3K navigation and 27.5K SDR task), R2R~\cite{anderson2018room2room} has 21.5K navigation instructions, REVERIE has 21.7K instructions, ALFRED~\cite{ALFRED20} has 25.7K language directives describing 8K demonstrations, and CVDN~\cite{thomason:corl19} dataset with 7.4K NDH instances and 2K navigation dialogues.

\noindent\textbf{Linguistic Properties.}
From our dataset, we randomly sampled 50 instructions for manual analysis. A unique linguistic property found in our sample is 3D discrete referring expressions which utilize 3D depth to guide the agent; implying that the combined navigation and assembly task requires that agents posses a full understanding of object relations in a 3D environment. Our analysis showed other linguistic properties, such as frequent directional references, ego and allocentric spatial relations, temporal conditions, and sequencing (see Appendix~\ref{app:ling_prop} for the details and examples).

\noindent\textbf{Dataset Statistics.}
Figure \ref{fig:itempie} shows that the most frequently occurring words in our dataset. These words are primarily directional or spatial relations. This implies that agents should be able to understand the concept of direction and the spatial relations between objects, especially as they change with movement. Table \ref{tbl:gt_instruct_stats} and Figure \ref{fig:wordcounts} show that navigation tends to have longer instructions and path lengths. Assembly occurs in a smaller environment, requiring agents to focus less on understanding paths than in navigation and more on understanding the 3D spatial relations of objects from the limited egocentric viewpoint.

\begin{table}[t]
\begin{center}
\resizebox{0.9\columnwidth}{!}{
 \begin{tabular}{|l|c|c|c|c|}
  \hline
 \multirow{2}{*}{Length} &\multicolumn{2}{c|}{Navigation} & \multicolumn{2}{c|}{Assembly}\\
 \cline{2-5} 
 & max & avg. & max & avg. \\
 \hline
 
    Instruction 
    & 147 & 47.99 & 90 & 20.99 \\
    \hline
    
    Path 
    & 156 & 48.14 & 8 & 3.32 \\
    \hline

    Action Sequence
    & 224 & 75.78 & 34 & 13.68 \\
    \hline 
\end{tabular}
}
\end{center}
\vspace{-10pt}
\caption{Lengths of the instructions (in words), paths, and action sequences for both turns across all subsections in the city. \label{tbl:gt_instruct_stats}
} 
\vspace{-5pt}
\end{table}

\section{Models}

We train an integrated Vision-and-Language model as a good starting point baseline for our task. To verify that our dataset is not biased towards some specific factors, we trained ablated and random walk models and evaluated them on the dataset.

\begin{figure}[t]
    \centering
    \includegraphics[width=0.98\columnwidth]{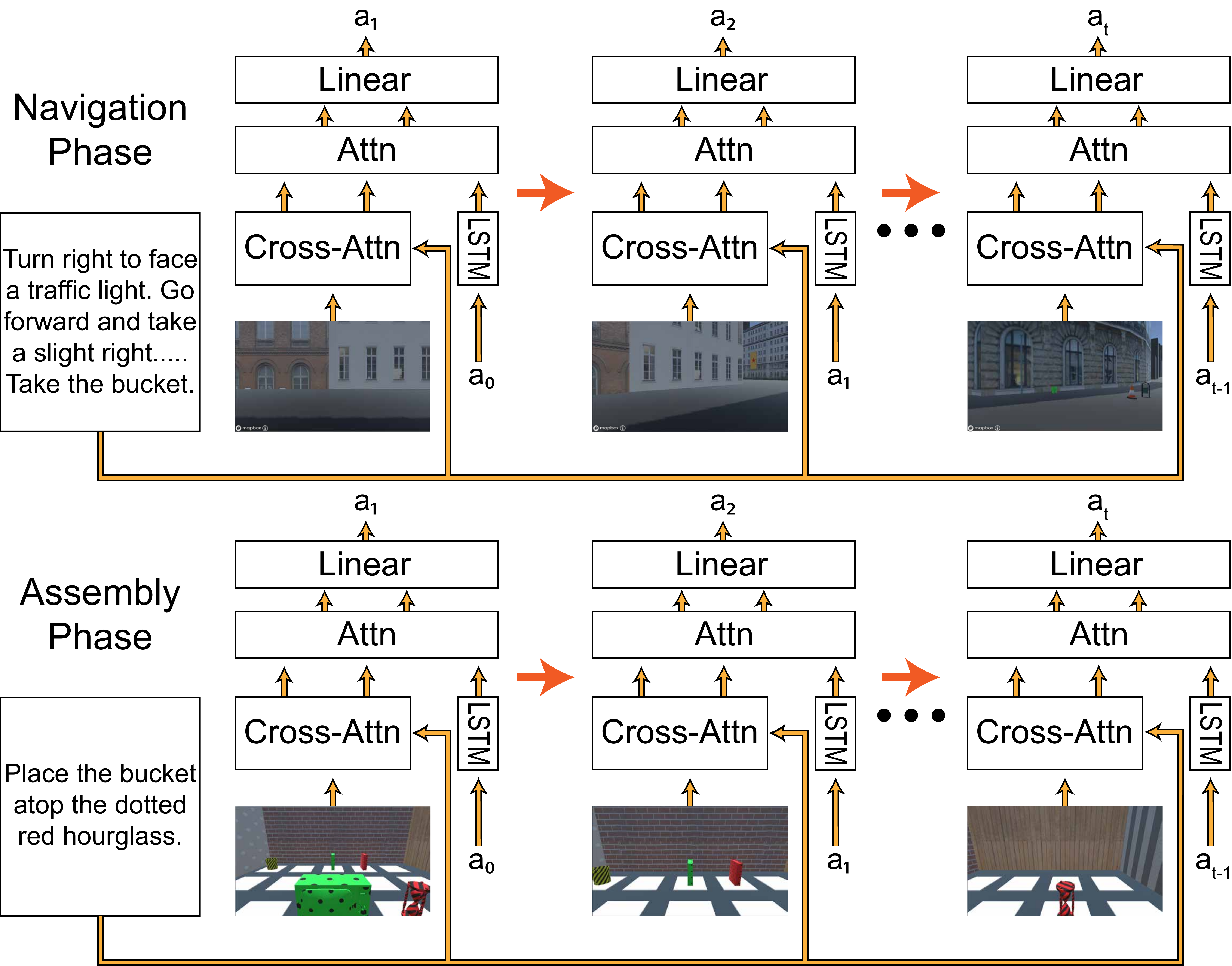}
    \vspace{-4pt}
    \caption{Vision-and-Language model: environment visual features, instruction language features, and action features are aligned to generate the next action.}
    \label{fig:modeldiagram}
    \vspace{-5pt}
\end{figure}

\noindent\textbf{Vision-and-Language Baseline.} This model uses vision and NL instruction features together to predict the next actions (Figure \ref{fig:modeldiagram}). We implement each module for navigation/assembly phases as:
\begin{align}
    L = \textrm{Emb}_L(&\textrm{Inst.}),
    \;\;\tilde{a}_t = \textrm{Emb}_A(a_t)\\
    V_{t} = \textrm{Enc}_V(&\textrm{Img}_{t}),\;\;
    \tilde{L} = \textrm{Enc}_L(L)\\
        h_{t} =&~\textrm{LSTM}(\tilde{a}_{t-1}, h_{t-1})\\
    \hat{V}_{t},\hat{L}_{t} =&~\textrm{Cross-Attn}(V_{t},\tilde{L})\\
    v_t =\textrm{Attn}(h_{t},& \hat{V}_t), \;\; l_t =\textrm{Attn}(h_{t},\hat{L}_{t})\\
    \textrm{logit}_{a_t} =\textrm{Linear}(&v_t, l_t), \;\; a_t = \textrm{max}(\textrm{logit}_{a_t})
\end{align}
where \(\textrm{Img}_{t}\) is the view of an agent at time step \(t\), Inst. is natural language instructions given to the agent, and \(a_{t}\) is an action at time step \(t\). Instructions and actions are embedded via \(\textrm{Emb}_L\) and \(\textrm{Emb}_A\), respectively. We use ResNet \cite{he2016deep} for the visual encoder, \(\textrm{Enc}_V\), to obtain visual features, \(V_t \in \mathbb{R}^{w\times w \times d_v}\), and LSTM \cite{hochreiter1997long} for the instruction encoder, \(\textrm{Enc}_L\), to obtain instruction features, \(\tilde{L} \in \mathbb{R}^{l \times d_l}\). We employ the bidirectional attention mechanism \cite{seo2016bidirectional} for the cross attention \(\textrm{Cross-Attn}\) to align the visual and instruction features, and use the general attention \(\textrm{Attn}\) to align the action feature and each of fused visual and instruction features. See Appendix~\ref{app:models} for the detailed descriptions of \(\textrm{Cross-Attn}\) and \(\textrm{Attn}\) modules.

We train the model with the teacher-forcing approach \cite{lamb2016professor} and cross entropy loss: $p_t(a_t) = \textrm{softmax}(\textrm{logit}_{a_t}); \,\,
\mathcal{L} = -\sum_t{\log{p_t(a_t^{*})}}$, where \(a_t^{*}\) is ground truth action at time step \(t\).

\noindent\textbf{Vision/Language only Baseline.}
To check the uni-modality bias, we evaluate vision and language only baselines on our dataset. These exploit only single modality (visual or language) to predict the appropriate next action. To be specific, they use the same architecture as the Vision-and-Language baseline except the \(\textrm{Cross-Attn}\) module.

\noindent\textbf{Random Walk.}
Agents take a random action at each time step without considering instruction and environment information. 

\noindent\textbf{Shortest Path.}
This baseline simulates an agent that follows the shortest path provided by A* algorithm \cite{hart1968Astar} to show that the GT paths are optimal in terms of trajectory distances.

\begin{table*}[t]
\begin{center}
\resizebox{2.0\columnwidth}{!}{
 \begin{tabular}{|c|c|c|c|c|c|c|c||c|c|c|c|c|c|c|}
  \hline
  \multirow{4}{*}{Model} & \multicolumn{7}{c||}{Val Seen} & \multicolumn{7}{c|}{Val Unseen} \\
  \cline{2-15} 
  &\multicolumn{5}{c|}{Navigation} & \multicolumn{2}{c||}{Assembly}&\multicolumn{5}{c|}{Navigation} & \multicolumn{2}{c|}{Assembly}\\
 \cline{2-15} 
  &\multirow{2}{*}{nDTW} & \multicolumn{4}{c|}{CTC}&

 \multirow{2}{*}{rPOD} &  \multirow{2}{*}{PTC}&\multirow{2}{*}{nDTW} & \multicolumn{4}{c|}{CTC}&

 \multirow{2}{*}{rPOD} &  \multirow{2}{*}{PTC}\\
 \cline{3-6} 
 \cline{10-13}
  &  & k=0 & k=3 & k=5 & k=7  &  & &  & k=0 & k=3 & k=5 & k=7  &  & \\
 \hline

    V/L   &0.135& 0.000 & 0.098 &0.149  & 0.200  & 0.058 &0.044
    & 0.109& 0.000&0.062 & 0.108 & 0.153&  0.036 &0.028\\
    \hline

    V/O  &0.055 & 0.000 & 0.043 & 0.062 & 0.087 & 0.008 &0.001
    & 0.043& 0.000 & 0.031 & 0.057 & 0.085 & 0.007 &0.002\\
    \hline

    L/O   & 0.110& 0.000 & 0.044 &0.095  & 0.147 & 0.023 & 0.017
    &0.105 & 0.000 &0.029  & 0.068 & 0.126 & 0.017 & 0.013\\
    \hline

    R/W   &0.045& 0.000 & 0.030 & 0.054 & 0.092 & 0.005   &0.001
    & 0.045& 0.000 &0.024 &0.043  & 0.075 & 0.005 & 0.001 \\
    \hline 
    S/P   &1.000 & - & 1.000 & 1.000 & 1.000 & -  & - 
    & 1.000& - & 1.000 & 1.000 & 1.000 & -  & - \\
    \hline 
    \hline
     H/W   &0.671 &1.000  &1.000  & 1.000 & 1.000 & 0.879 & 0.861
    &0.670 & 1.000 &1.000  & 1.000 & 1.000 & 0.869 &0.856\\
    \hline
    
\end{tabular}
}
\end{center}
\vspace{-10pt}
\caption{Performance of baselines and humans on the metrics for the Val-Seen/Unseen splits. Overall, there is large human-model performance gap, indicating our \dataname{} task is very challenging (V/L:Vision-and-Language, V/O:Vision-Only, L/O:Language-Only, R/W:Random-Walk, S/P:Shortest Path, H/W:Human-Workers). \label{tbl:main_result} 
} 
\vspace{-5pt}
\end{table*}

\section{Experiments}
We split the dataset into train/val-seen/val-unseen/test-unseen. We assign the city sub-sections 1 to 5 to train and val-seen, sub-section 6 to val-unseen, and section 7 to test-unseen splits. We randomly split data from sub-sections 1 to 5 into 80/20 ratio to get train and val-seen splits, respectively. Thus, the final number of task samples for each split is 4,267/1,065/1,155/1,205 (total: 17,068/4,260/4,620/4,820). 
The Stage 1 workers are equally distributed across the city sub-sections, so the dataset splits are not biased toward specific workers. We also keep the separate 2 sections (i.e., section 6 and 7) for the unseen dataset following \citet{anderson2018room2room}, which allows the evaluation of the models' ability to generalize in new environments.
Note that for agents to proceed to the next phase, we allow them to pick up the closest target object (in the navigation phase) or place collected object at the closest location (in the assembly phase) when they do not perform the required actions.
Training Details: We use 128 as hidden size. For word and action embedding sizes, we use 300 and 64, respectively. We use Adam \cite{kingma2014adam} as the
optimizer and set the learning rate to 0.001 (see Appendix~\ref{app:train_detail} for details).

\begin{table}[t]
\begin{center}
\resizebox{0.999\columnwidth}{!}{
 \begin{tabular}{|c|c|c|c|c|c|c|c|}
  \hline
  \multirow{4}{*}{Model} & \multicolumn{7}{c|}{Test Unseen} \\
  \cline{2-8} 
  &\multicolumn{5}{c|}{Navigation} & \multicolumn{2}{c|}{Assembly}\\
 \cline{2-8} 
  &\multirow{2}{*}{nDTW} & \multicolumn{4}{c|}{CTC}&

 \multirow{2}{*}{rPOD} &  \multirow{2}{*}{PTC}\\
 \cline{3-6} 
  &  & k=0 & k=3 & k=5 & k=7  &  & \\
 \hline

    V/L   &0.114 & 0.000 &0.082  & 0.122 &  0.168& 0.047 &0.035\\
    \hline
    \hline
     H/W   &0.664 & 1.000 & 1.000 & 1.000 &1.000  & 0.884 &0.873\\
    \hline
    
    H/E  &0.806 & 1.000 & 1.000 &  1.000& 1.000 & 0.992 & 0.990\\
    \hline

\end{tabular}
}
\end{center}
\vspace{-10pt}
\caption{The Vision-and-Language (V/L) baseline and Human performance on Test-Unseen split (H/W:Human-Workers, H/E:Human-Expert). \label{tbl:main_test_unseen} } 
\vspace{-10pt}
\end{table}

\section{Results and Analysis \label{sec:results}}
As shown in Table \ref{tbl:main_result}, overall, there is large human-model performance gap, indicating that our \dataname{} task is very challenging and there is much room for model improvement. Performance in the navigation and assembly phases are directly related. If perfect performance is assumed in the navigation phase, rPOD and PTC are higher than if there were low CTC-k scores in navigation (e.g., 0.382 vs. 0.044 for PTC of the Vision-and-Language model on val-seen: see Appendix~\ref{app:results} for the comparison). This scoring behavior demonstrates that phases in our \dataname{} task are interweaved. Also, comparing scores from turn 1 and 2, all turn 2 scores are lower than their turn 1 counterparts (e.g., 0.222 vs. 0.049 nDTW of the Vision-and-Language model on val-seen split; see Appendix~\ref{app:results} for the detailed turn-wise results). This shows that the performance of the previous turn strongly affects the next turn's result. Note that to relax the difficultly of the task, we consider CTC-3 (instead of CTC-0; see Section \ref{sec:metrics}) as successfully picking up the target object and then we calculate the assembly metrics under this assumption. If this was not done, then almost all the metrics across assembly would be nearly zero.

\subsection{Model Ablations}

\noindent\textbf{Vision/Language Only Baseline.}
As shown in Table \ref{tbl:main_result}, our Vision-and-Language baseline shows better performance over both vision-only and language-only models, implying our dataset is not biased to a single modality and requires multimodal understanding to get high scores. 

\noindent\textbf{Random Walk.}
The Random-Walk baseline shows poor performance on our task, implying that the task cannot be solved through random chance.

\noindent\textbf{Human Evaluation.}
We conducted human evaluations with workers (Table \ref{tbl:main_result}, \ref{tbl:main_test_unseen}) as well as an expert (Table \ref{tbl:main_test_unseen}). For workers' evaluations, we averaged all the workers' scores for the verified dataset (from Stage 2: verification/following, see Sec.~\ref{sec:datacollection}). For expert evaluation, we took 50 random samples from test-unseen and asked our simulator developer to blindly complete the task. Both workers and the expert show very high performance on our task (0.66 nDTW and 0.87 PTC for workers; 0.81 nDTW and 0.99 PTC for expert), demonstrating a large model-human performance gap and allowing much room for further improvements by the community on our challenging \dataname{} dataset.

\subsection{Output Examples}
\begin{figure}[t]
    \centering
    \includegraphics[width=0.99\columnwidth]{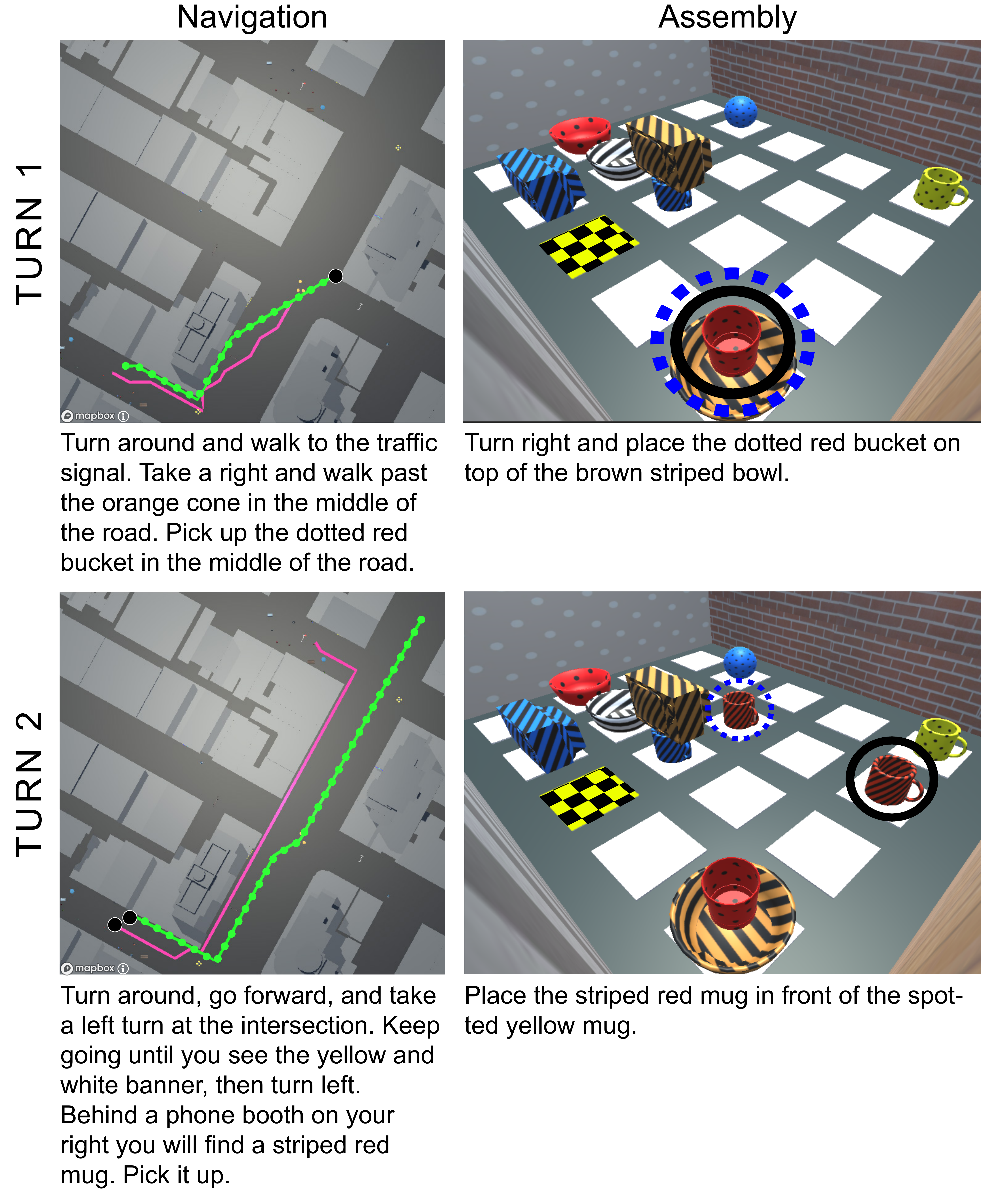}
    \vspace{-5pt}
    \caption{Visual demonstrations by our model in navigation and assembly phases (top-down view for illustration). GT navigation paths are solid pink lines and model's paths are dotted green lines (start = black dot). GT assembly target location is solid black circle and model's target object placement is dashed blue circle (start = checkered yellow tile, agent facing brick wall).}
    \label{fig:path_gen}
    \vspace{-10pt}
\end{figure}

As shown in an output example in Figure \ref{fig:path_gen}, our model navigates quite well and reaches very close to the target in the 1st turn and then places the target object in the right place in the assembly phase. However, in the 2nd turn, our model fails to find the ``striped red mug" by missing the left turn around the ``yellow and white banner". In the next assembling phase, the model cannot identify the exact location (``in front of the spotted yellow mug") to place the collected object (assuming the model picked up the correct object in the previous phase) possibly being distracted by another mug and misunderstanding the spatial relation. See Appendix~\ref{app:output_examples} for more output examples.

\section{Multilingual Setup}
We have also expanded our dataset now to include Hindi instructions. In the future, we are hoping to expand further to other languages.

\section{Conclusion}
We introduced \dataname{}, a new joint navigation+assembling instruction following task in which agents collect target objects in a large realistic outdoor city environment and arrange them in a dynamic grid space from an egocentric view. We collected a challenging dataset (in English and now also Hindi) via a 3D synthetic simulator with diverse object referring expressions, environments, and visuospatial relationships. 
We also provided several baseline models which have a large performance gap compared to humans, implying substantial room for improvements by future work.

\section*{Acknowledgments}
We thank the reviewers for their helpful comments. This work was supported by NSF Award 1840131, ARO-YIP Award W911NF-18-1-0336, DARPA MCS Grant N66001-19-2-4031, and a Google Foucused Award. The views contained in this article are those of the authors and not of the funding agency.

\bibliography{anthology,emnlp2020}
\bibliographystyle{acl_natbib}

\appendix
\section*{Appendices}

\section{Task and Metrics}
\begin{figure}[t]
    \centering
    \includegraphics[width=0.75\columnwidth]{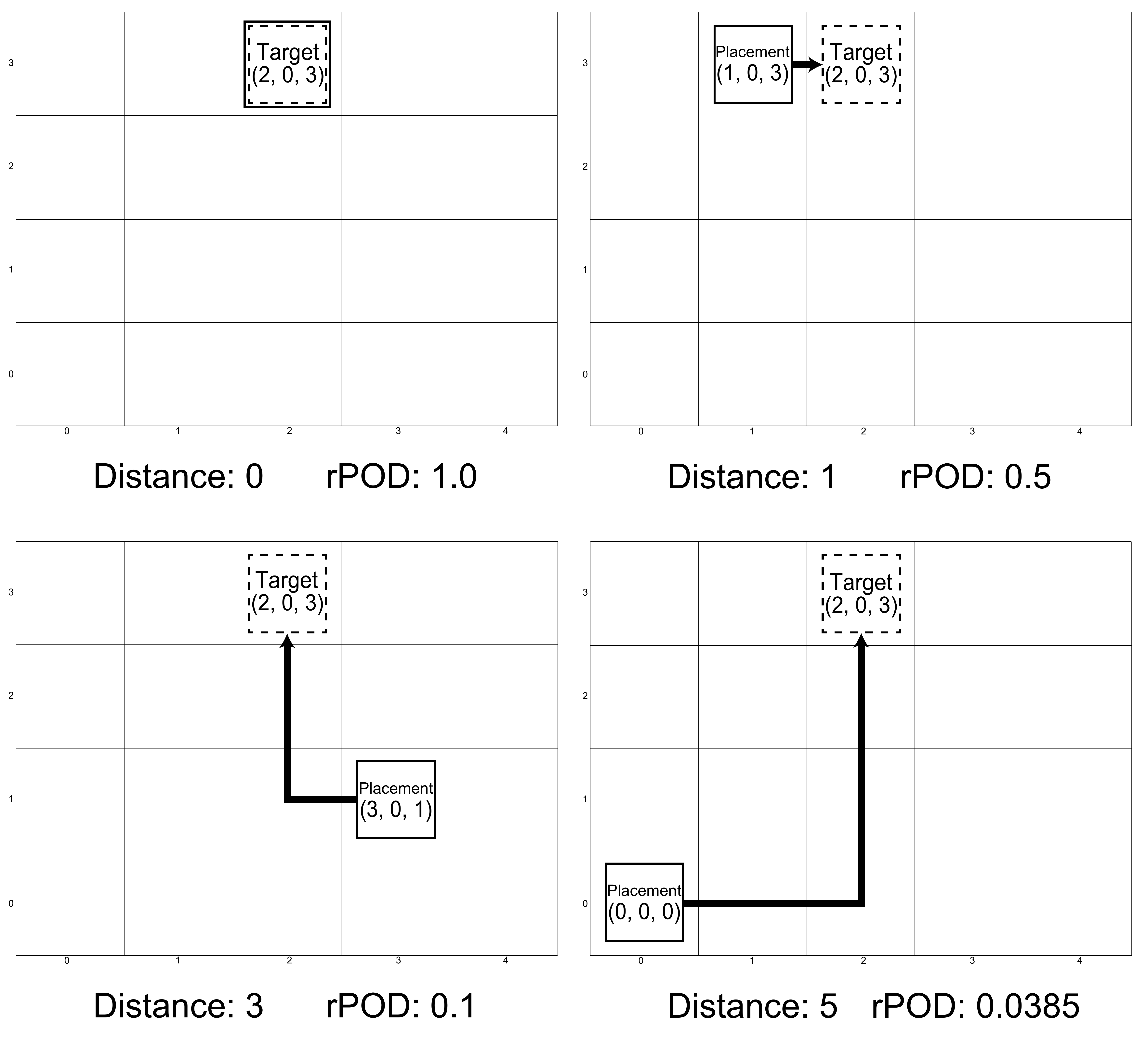}
    \caption{Distance and rPOD metric: as the Manhattan distance between target and agent placement locations increases, the rPOD score decreases exponentially.}
    \label{fig:rpod}
\end{figure}

As shown in Figure \ref{fig:rpod}, the score of rPOD is decreased according to the placement error (the Manhattan distance) exponentially. Thus, to score high in the rPOD metric, agents should place the target objects as close to the target place as possible.

\section{Dataset}
To support the \dataname{} task, we collected a dataset. Our dataset is based on a large dynamic outdoor environment from which diverse instructions with interesting linguistic properties are derived. 
\subsection{Data Collection\label{app:data_collect}}
\begin{figure*}[t]
\begin{subfigure}{.5\textwidth}
    \centering
    \includegraphics[width=.75\columnwidth]{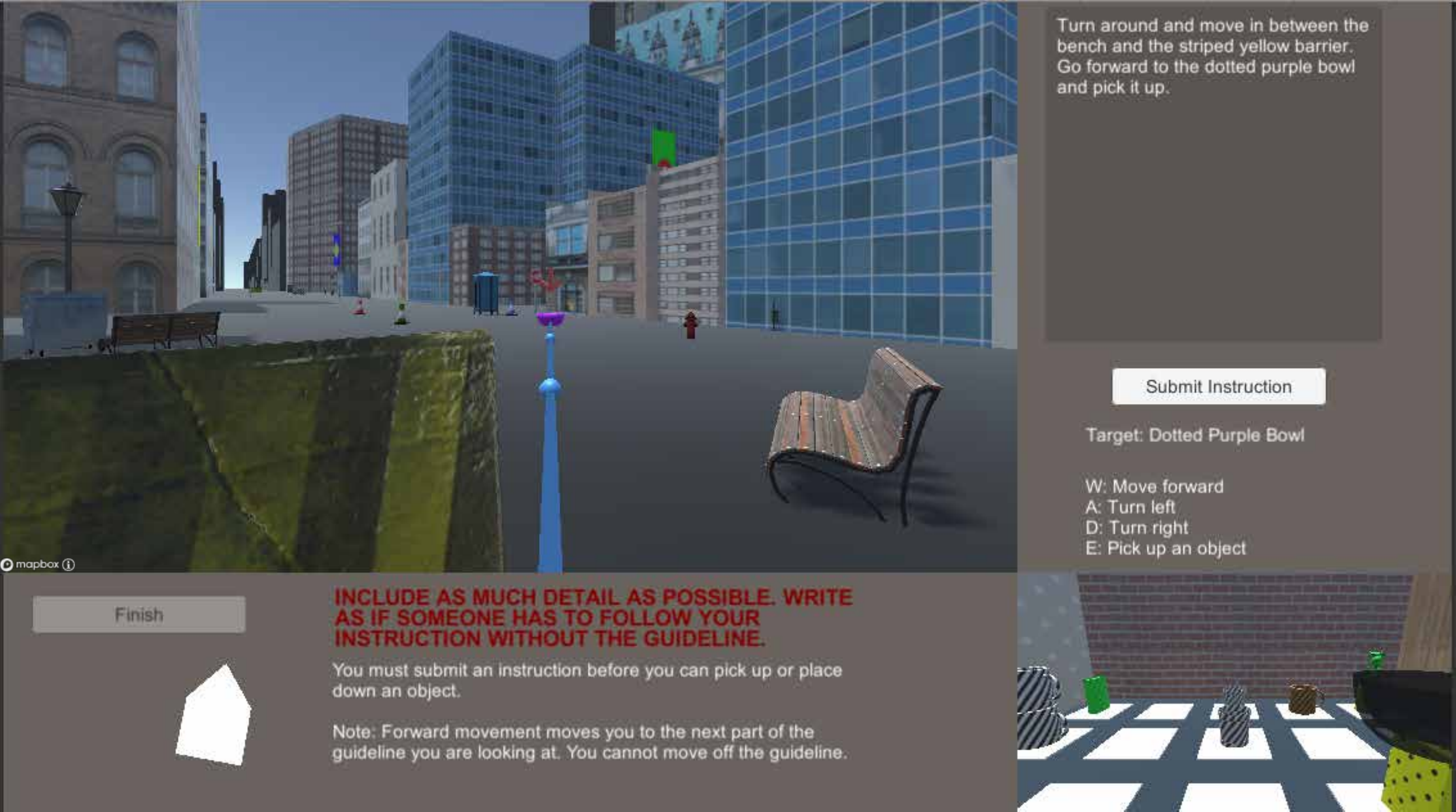}
    \caption{Navigation phase in stage 1.}
    \label{fig:stage1_navigation}
\end{subfigure}
\begin{subfigure}{.5\textwidth}
    \centering
    \includegraphics[width=.75\columnwidth]{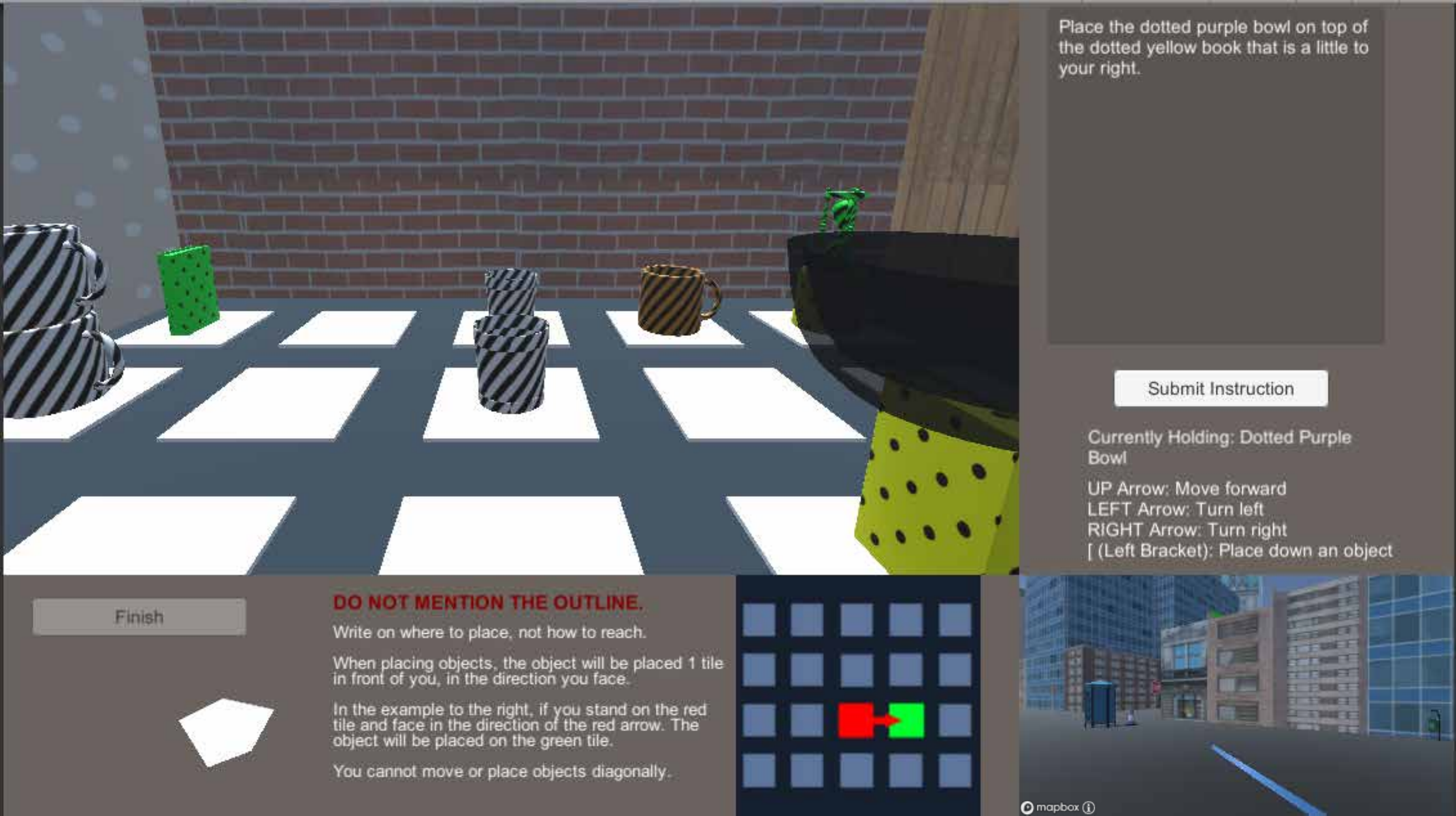}
    \caption{Assembly phase in stage 1.}
    \label{fig:stage1_assembly}
\end{subfigure}
\newline
\begin{subfigure}{.5\textwidth}
    \centering
    \includegraphics[width=.75\columnwidth]{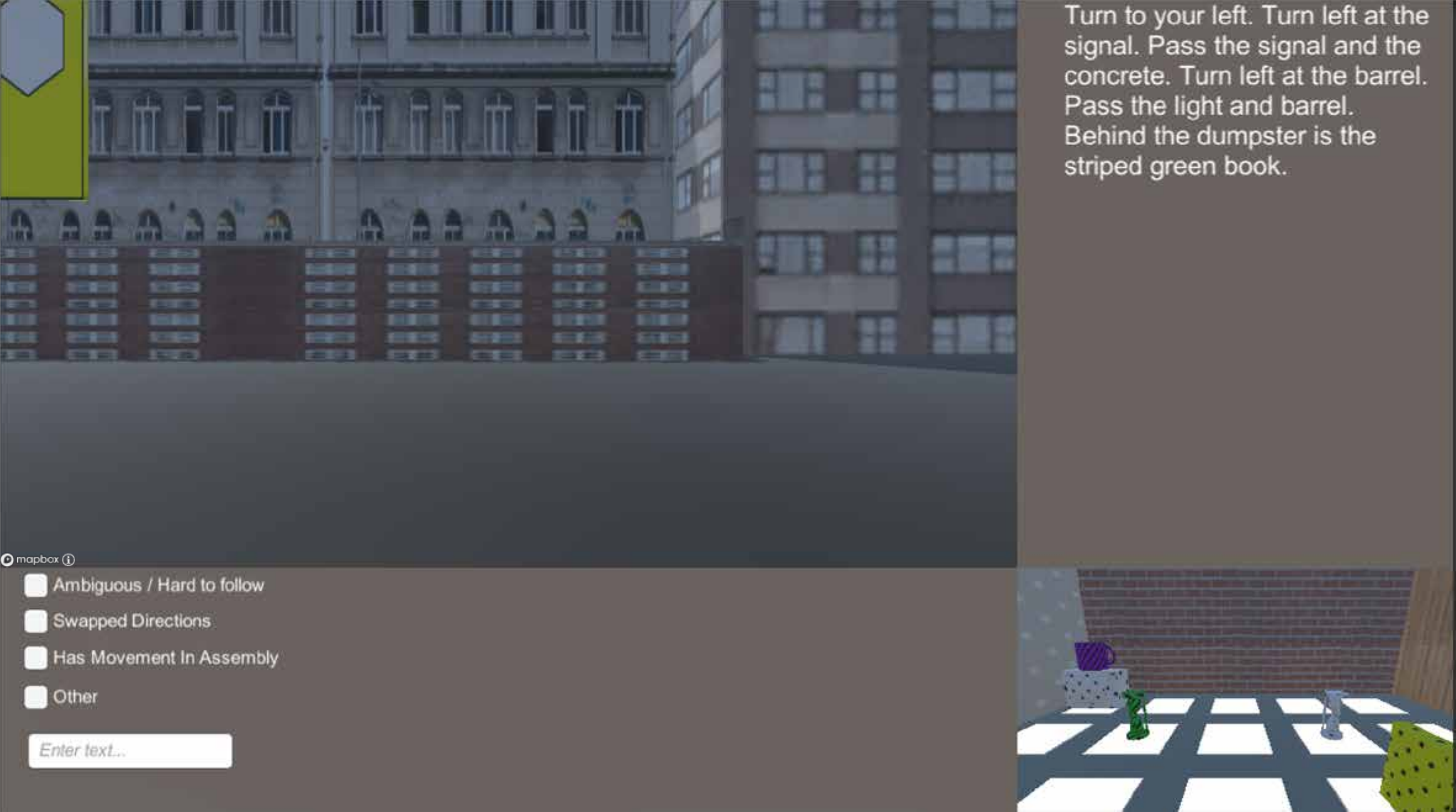}
    \caption{Navigation phase in stage 2.}
    \label{fig:stage2_navigation}
\end{subfigure}
\begin{subfigure}{.5\textwidth}
    \centering
    \includegraphics[width=.75\columnwidth]{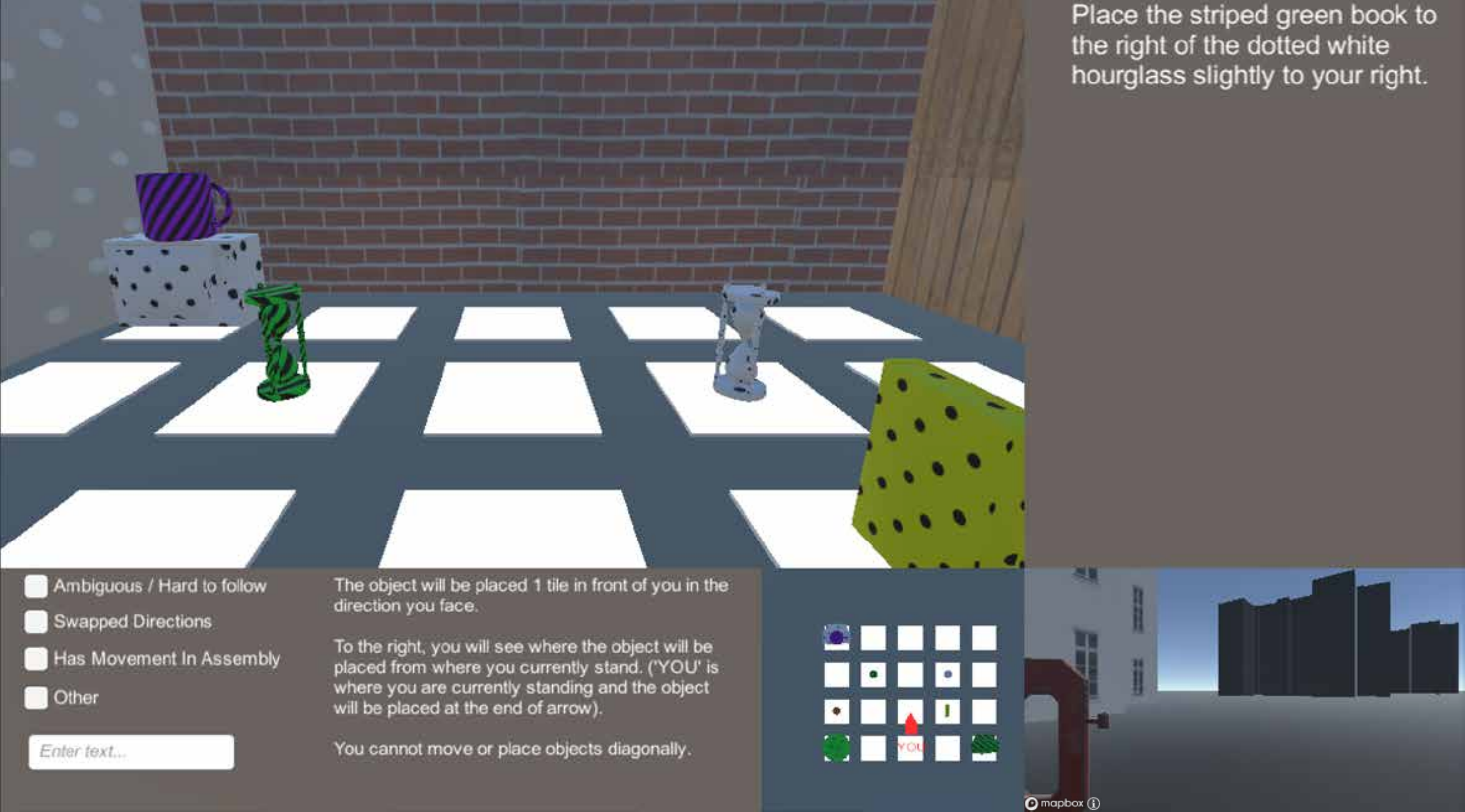}
    \caption{Assembly phase in stage 2.}
    \label{fig:stage2_assembly}
\end{subfigure}
\caption{Simulation Interfaces of the Stage 1 (upper), Stage 2 (lower) showing separate examples, navigation phase (left), and assembly phase (right) of the data collection. (a) Workers are initially shown the navigation phase interface and must follow the blue navigation line to the target objects and write instructions as they go. (b) Workers are moved into assembly and must make assembly instructions guided by the highlighted (transparent black) objects. (c) Workers are provided with the navigation instructions and must find the target objects identified by the instructions. (d) Workers are provided with the assembly instructions and must place the collected object at the target position identified by the instructions.}
\label{fig:simulation_interface}
\end{figure*}

\paragraph{Route Generation.} 
The ground truth trajectories is determined by the A* shortest path algorithm \cite{hart1968Astar}. Using the shortest path algorithm allows the resulting Ground Truth (GT) path to be straightforward and reach the target while avoiding going to unnecessary places. The blue navigation guideline provided to the Stage 1 workers is a mimic of this GT path (Figure \ref{fig:stage1_navigation}).

\begin{figure}[t]
    \centering
    \includegraphics[width=0.8\columnwidth]{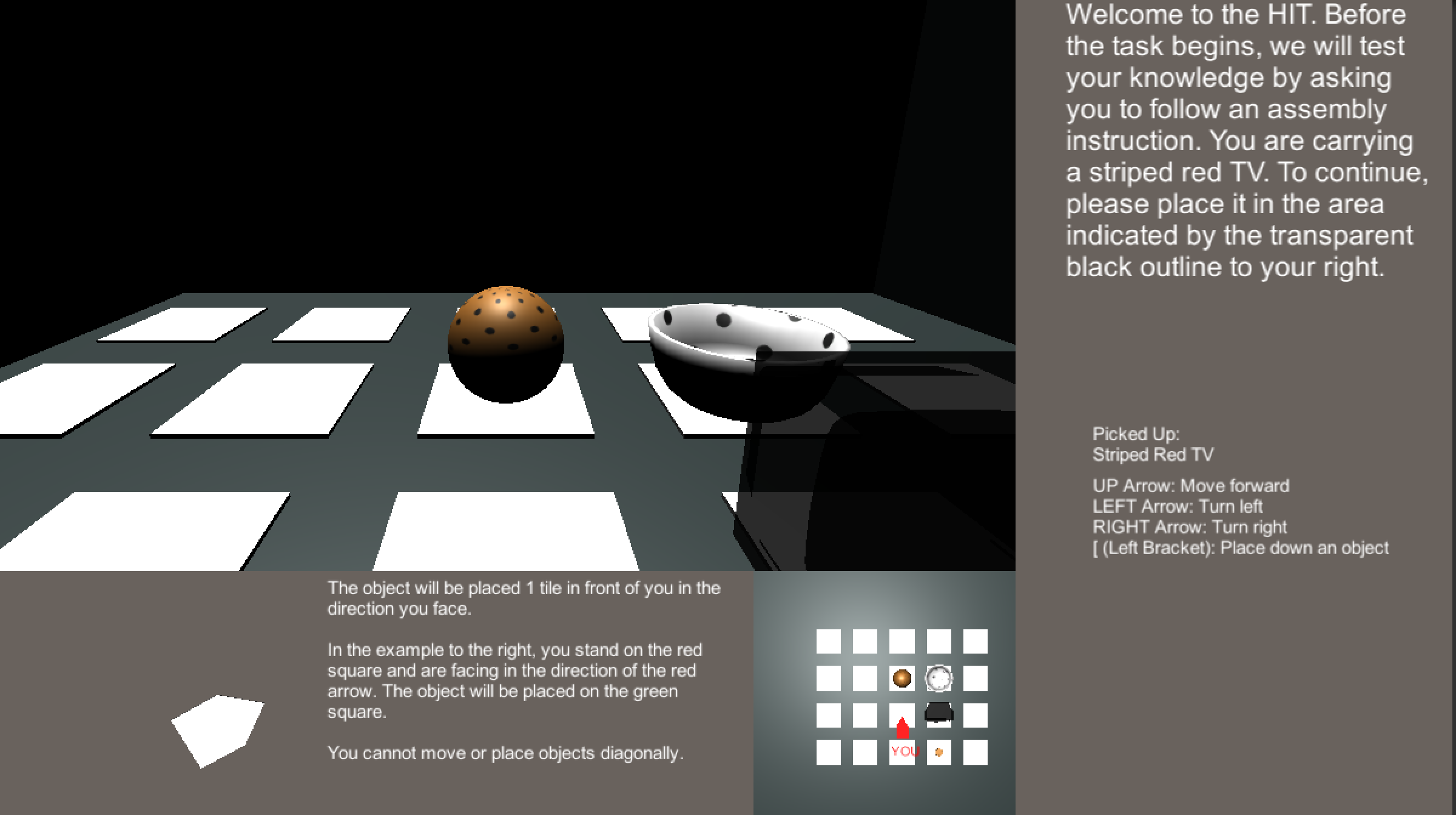}
    \caption{Illustration of the assembly phase test before the start of stage 2.}
    \label{fig:stage2_assembly_test}
\end{figure}

\begin{figure}[t]
    \centering
    \includegraphics[width=0.65\columnwidth]{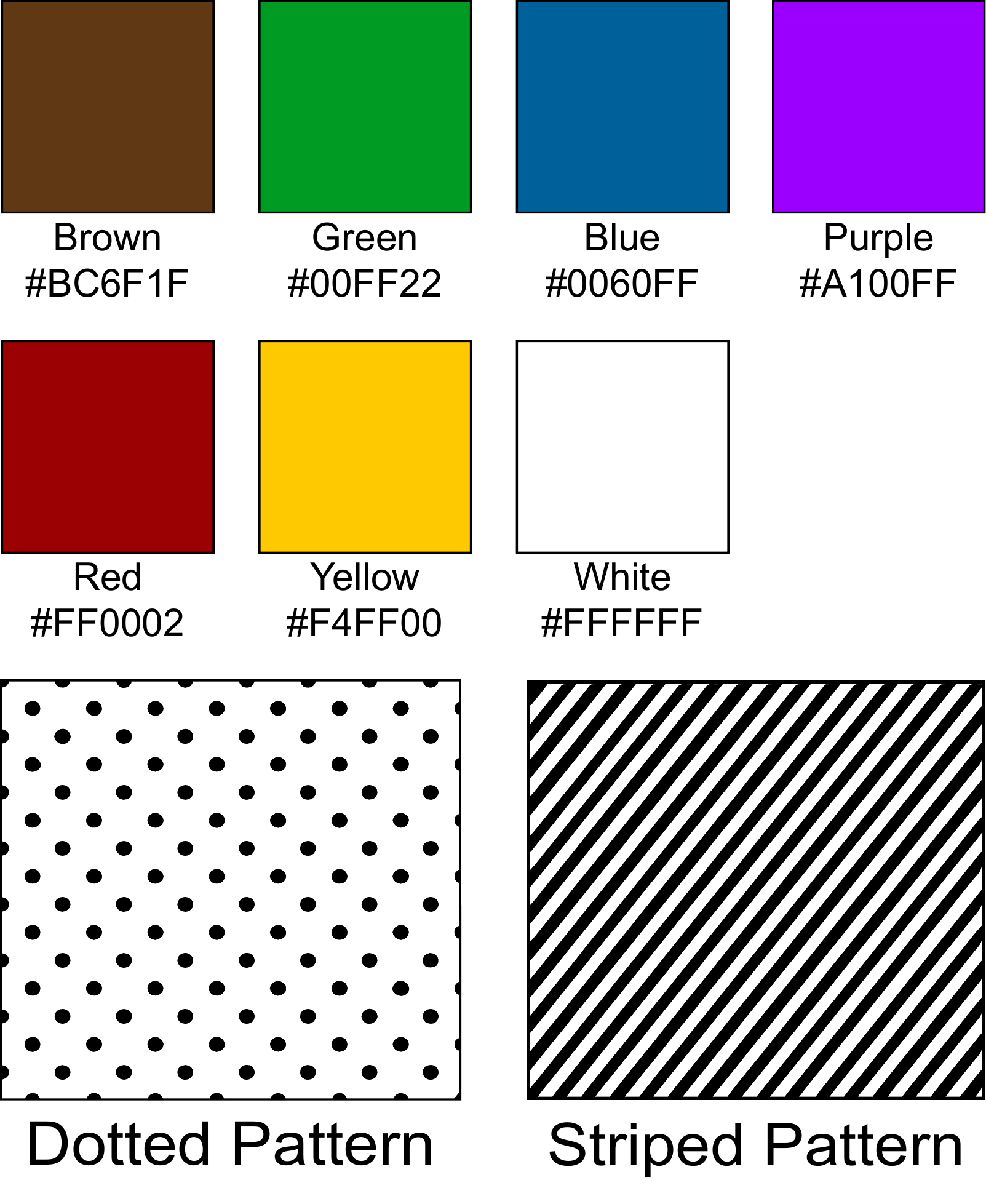}
    \caption{Illustration of the colors and patterns that collectable and distracter objects can have.}
    \label{fig:colortexture_graphic}
\end{figure}

\begin{figure}[t]
    \centering
    \includegraphics[width=0.5\columnwidth]{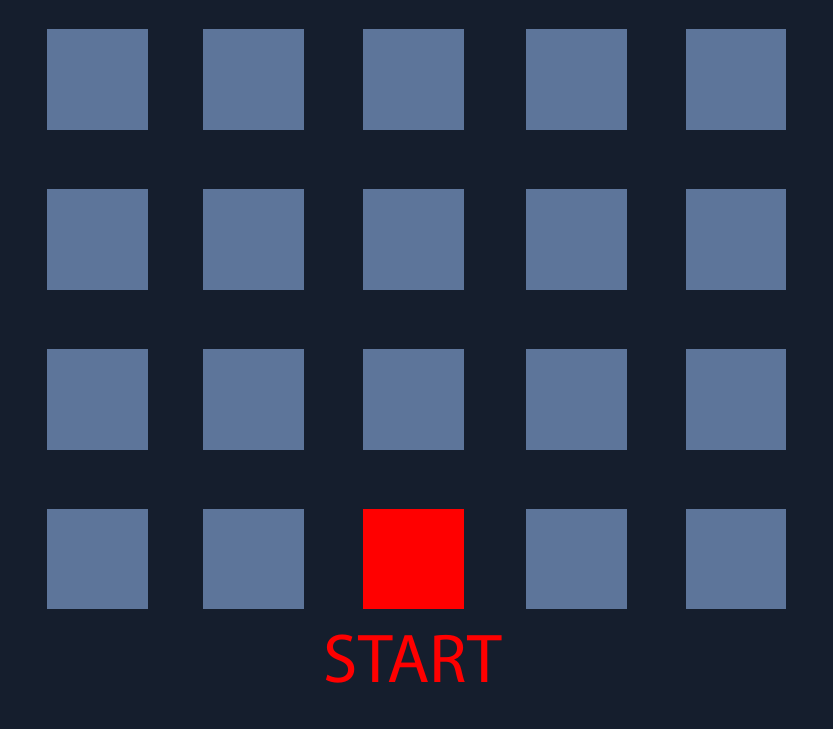}
    \caption{Illustration of the assembly grid with the starting position marked.}
    \label{fig:grid_starting_position}
\end{figure}

\begin{figure*}[t]
    \centering
    \includegraphics[width=1.95\columnwidth]{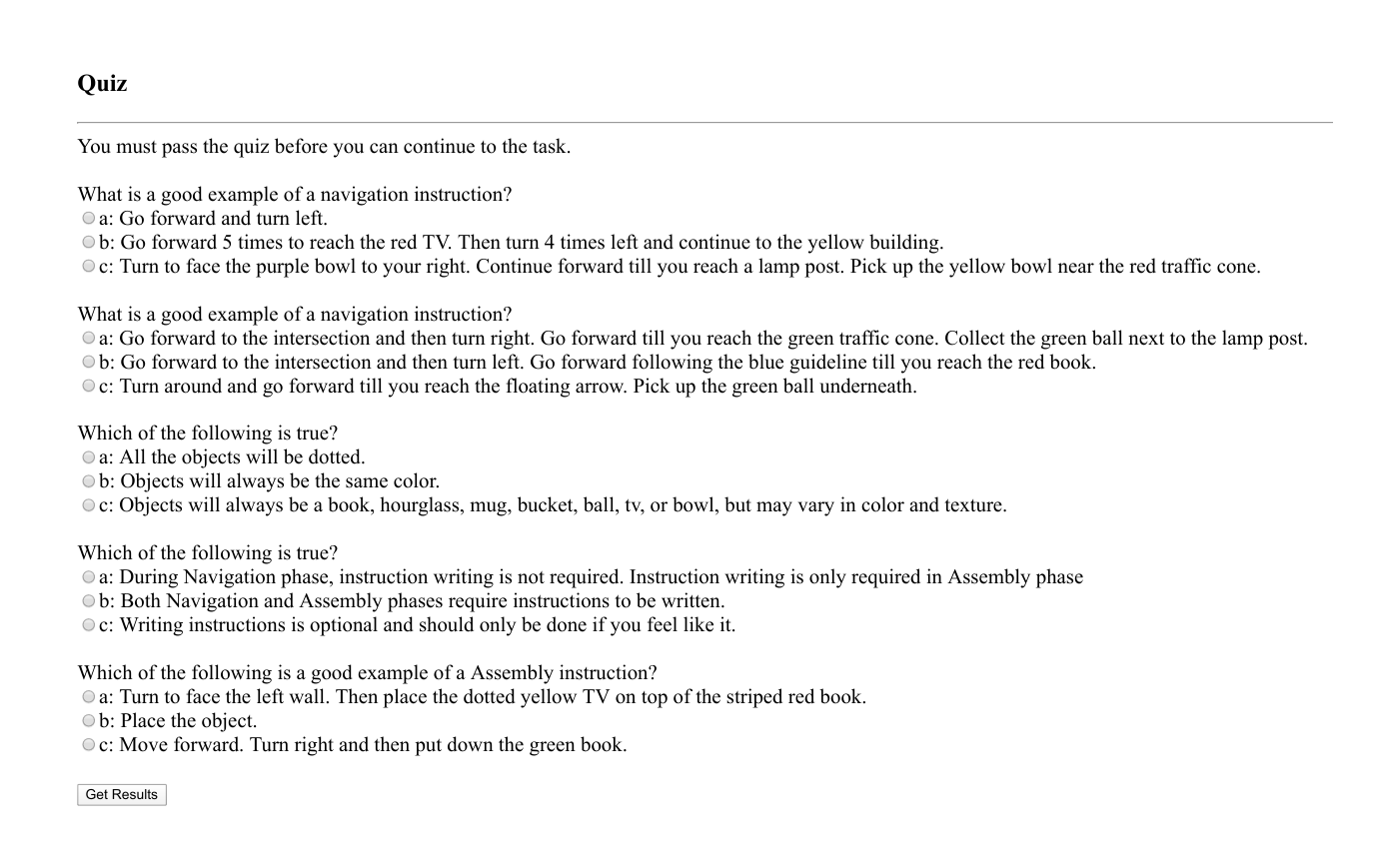}
    \caption{Screening test that is required to be taken prior to starting Stage 1.}
    \label{fig:quiz_stage1}
\end{figure*}

\begin{figure*}[t]
    \centering
    \includegraphics[width=1.95\columnwidth]{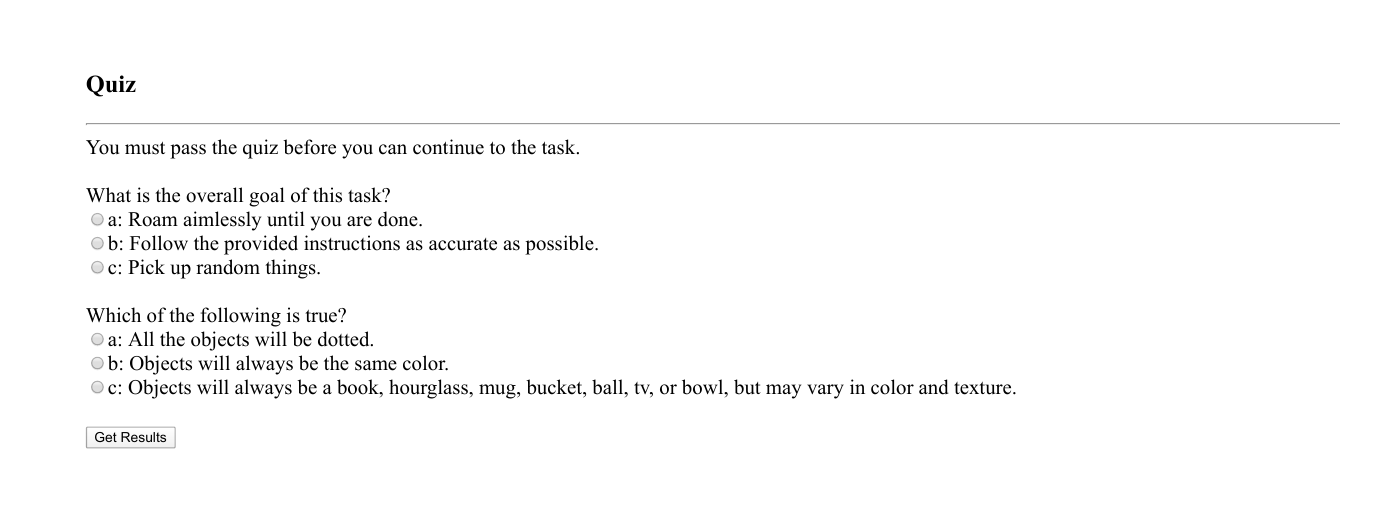}
    \caption{Screening test that is required to be taken prior to starting Stage 2.}
    \label{fig:quiz_stage2}
\end{figure*}

\paragraph{Qualification Tests.} When placing an object in the assembly phase, the item is placed 1 space in front of where the agent stands. To ensure that the workers who will be following instructions in Stage 2 fully understood this concept, at the start of Stage 2, they were presented with a small test (Figure \ref{fig:stage2_assembly_test}) that would show them how to correctly move and place objects and required that they demonstrate that they could do so. Both Stage 1 and 2 workers were also required to pass a short screening test before they could begin their respective tasks. The tests are shown in Figure \ref{fig:quiz_stage1} (Stage 1) and Figure \ref{fig:quiz_stage2} (Stage 2).

\paragraph{Worker Bonus Criteria and Rates.} For Stage 1 workers who did the instruction writing task correctly \{5, 20, 50\} times, a bonus of \{\$0.10, \$0.90, \$4.00\} respectively was awarded. Stage 1 workers were also provided a \$0.10 bonus for every instruction they wrote that was able to successfully pass Stage 2 verification with high nDTW and perfect assembling scores.

\paragraph{Instruction Rules and Guidelines.}
Rules and guidelines were put into place to help ensure that instructions written by the Stage 1 workers were high quality and written with as few errors as possible. Particularly, the guidelines serve to prevent the workers from using other elements of the UI or tools we provided, such as the blue navigation line or guiding arrow (see Figure \ref{fig:simulation_interface}) and other elements that were not part of the true environment in their instructions.
\begin{itemize}[leftmargin=*,noitemsep]
    \item Instructions must be written relative to objects and the environment and not contain exact counts of movements (e.g., ``Go forward 10 times and then turn left 2 times" is bad).
    \item Instructions must be clear, concise, and descriptive.
    \item Do not write more than the text-field can hold.
    \item At the end of writing an instruction for the navigation phase be sure to include something similar to ``pick up" or ``collect" the object.
    \item At end of writing an instruction for the assembly phase be sure to include something similar to ``place" or ``put" the object you collected before.
    \item Do not reference the navigation line, the blue balls on the navigation line, the floating arrows above the objects, or any of the interface elements when writing instructions.
    \item Do not reference any buildings that are a solid gray color.
    \item Do not reference the transparent black outline or the white grid tiles on the floor (Figure \ref{fig:grid_starting_position} and Figure \ref{fig:stage1_assembly}) during the assembly phase
    \item Do not write vague or potentially misleading instructions and do not create any instructions that reference previous instructions such as ``Go back to" or ``Return to".
    \item Avoid spelling and grammar mistakes.
    \item When writing instructions for the assembly phase, do not write movement instructions. Make sure to use object references (e.g, ``the red dotted ball").
\end{itemize}
During the navigation phase, the instruction writing worker cannot stray from the navigation line, ensuring that they collect the objects in the correct order. During the assembly phase, regardless of where the instruction worker places the collected object, it will move into the correct position (workers are not informed of this), ensuring that the objects are always in the correct formation for the next phase and future instructions do not become invalid. Additionally, we have implemented active quality checks which will prevent a worker from submitting their instructions if certain criteria is not met. If a worker is blocked by one of these checks, they will be shown which check failed so that they can easily correct the error.
\paragraph{General Active Quality Checks.}
\begin{itemize}[leftmargin=*,noitemsep]
    \item Each instruction must contain at least 6 words.
    \item Less than 40\% of the characters in the instructions can be spaces.
    \item The symbols (, [, ], ), \&, *, \^{}, \%. \$, \#, @, !, =, and + cannot be included.
    \item Single letter words other than "a" cannot be included.
    \item A single letter cannot be repeated 3 consecutive times. i.e ``sss".
    \item The same word cannot be repeated twice in a row.
    \item At least 40\% of the words in the instruction must be unique.
    \item The term ``key" cannot be included.
    \item The term ``step" cannot be included.
    \item The term ``time" cannot be included.
    \item The term ``go back" cannot be included.
    \item The term ``return" cannot be included.
    \item The term ``came" cannot be included.
    \item The term ``item" cannot be included.
\end{itemize}
\paragraph{Navigation Active Quality Checks.}
\begin{itemize}[leftmargin=*,noitemsep]
    \item If the ground truth path requires turning at the beginning of the path, the term ``turn" must be included.
    \item The term ``arrow" cannot be included.
\end{itemize}
\paragraph{Assembly Active Quality Checks.}
\begin{itemize}[leftmargin=*,noitemsep]
    \item The terms ``tile" or ``grid" cannot be included.
    \item The term ``space" cannot be included.
    \item The term ``go" cannot be included.
    \item The term ``corner" cannot be included.
    \item The term ``move" cannot be included.
    \item The black outline cannot be referenced.
\end{itemize}
\paragraph{Review Notifications.} It is possible for instructions to be written that can pass all automated checks and still be of poor quality. However, there is no quick and reliable way to automatically check if an instruction passes the tests but is still vague or misleading. Additional active checks could be added, however, in cases of ambiguity, more active checks would result in potentially good instructions being blocked. Instead of blocking submission, checks that could have been incorrectly triggered, would send a notification email, allowing us to take quick action by manually reviewing the instruction in question to see if the worker who created it needs feedback on writing better instructions.

\begin{table*}[t]
\begin{center}
\resizebox{1.95\columnwidth}{!}{
 \begin{tabular}{|c|c|c|c|c|c|c|c|}
  \hline
 Linguistic Property & 
Navigation Frequency &
 Assembly Frequency &
 Instruction Examples \\
 \hline
 \hline
   \multirow{2}{*}{Egocentric Spatial Relation} & \multirow{2}{*}{34\%}  & \multirow{2}{*}{34\%} & \multirow{2}{*}{\shortstack[l]{``...Go straight so the striped green bucket with\\ the red tv on top of it is to \textbf{your right}..."}}\\
   &&&\\
   \hline
   \multirow{2}{*}{Allocentric Spatial Relation} & \multirow{2}{*}{86\%} & \multirow{2}{*}{98\%} & \multirow{2}{*}{\shortstack[l]{``Place the dotted yellow bucket \textbf{on the left side of} the striped brown bowl."}}\\
   &&&\\
   \hline
    \multirow{2}{*}{Temporal Condition} & \multirow{2}{*}{64\%} & \multirow{2}{*}{2\%} & \multirow{2}{*}{\shortstack[l]{``...Continue to walk forward \textbf{until you reach an intersection}..."}}\\
   &&&\\
    \hline
   \multirow{2}{*}{Directional Reference} & \multirow{2}{*}{96\%} & \multirow{2}{*}{68\%} & \multirow{2}{*}{\shortstack[l]{``Make \textbf{a slight left} and walk forward stopping at the intersection."}}\\
   &&&\\
   \hline
   \multirow{2}{*}{Sequencing} & \multirow{2}{*}{66\%} & \multirow{2}{*}{58\%} & \multirow{2}{*}{\shortstack[l]{``...Go forward past the dotted yellow bucket and\\ past the lamp post near the blue phone booth..."}}\\
   &&&\\
   \hline
   \multirow{2}{*}{3D Discrete Referring Expressions} & \multirow{2}{*}{72\%} & \multirow{2}{*}{34\%} & \multirow{2}{*}{\shortstack[l]{``Put the striped blue book behind the dotted red mug."}}\\
   &&&\\
   \hline
\end{tabular}
}
\end{center}
\vspace{-10pt}
\caption{Linguistic properties and their frequencies found in within 50 randomly sampled instruction sets from the \dataname{} dataset.
\label{tbl:lang_analy}
}
\end{table*}

\subsection{Interface\label{app:interface}}
\paragraph{Stage 1: Instruction Writing.} The goal of this stage is to write instructions on how to navigate and place objects. The provided interface was designed to make this process easier for the workers completing the task. In both phases, the interface provides a arrow on the bottom left that will also point to the target destination and target location (depending on the active phase; navigation and assembly respectively.)
\begin{itemize}[leftmargin=*,noitemsep]
    \item \textbf{Navigation Phase:} (Figure \ref{fig:stage1_navigation}) The workers will follow the provided navigation line and as they follow it, write instructions on how to reach the destination. Additionally, the workers are provided with the controls and a few tips that they should keep in mind while completing the navigation phase. A small preview of the next phase (Assembly) is shown in the lower right.
    \item \textbf{Assembly Phase:} (Figure \ref{fig:stage1_assembly}) The interface is similar to that of the navigation phase interface. During this phase, the Assembly preview which previously occupied the lower right corner will come into focus, and the navigation phase preview is now occupying that space. In this phase, no navigation line is provided, as there is nowhere that cannot be seen from the starting position. The controls and tip information are updated with information about the assembly phase.
\end{itemize}
\paragraph{Stage 2: Instruction Following.} The goal of this stage is for the instructions written in the previous to be validated. Again, this interface was designed to make completing this task easier for the workers. Workers are also provided with some check boxes, which they can use to flag an instruction for certain issues so that we can more easily identify poor instructions.
\begin{itemize}[leftmargin=*,noitemsep]
    \item \textbf{Navigation Phase:} (Figure \ref{fig:stage2_navigation}) Workers are placed in an exact copy of the environment that a Stage 1 worker used, as well as given the instructions they wrote on how to accomplish the task, which are visible in the top right corner. This new worker is not provided the blue guideline and the indicating arrow, and must now navigate using the instructions alone.
    \item \textbf{Assembly Phase:} (Figure \ref{fig:stage2_assembly}) The worker is again shifted into the assembly room, but will no longer see the transparent outline that indicates where the object should be placed. They must instead rely on the instructions written by a Stage 1 worker. The worker is also provided a real-time diagram indicating where they will place the object given the position they currently stand. The object is always placed 1 space directly in front of the worker's location. The worker is also provided with some tips that might help them.
\end{itemize}

\section{Data Analysis}
\subsection{Linguistic Properties\label{app:ling_prop}} As shown in Table \ref{tbl:lang_analy}, our instruction sets have diverse linguistic features that make our task more challenging. Our \dataname{} task requires that the agent be able to understand and distinguish between both egocentric and allocentric spatial relations, necessitating that they comprehend the relation between entities in the environment according to their location and orientation. The instructions contain many directional words and phrases which require that agents utilize strong navigational skills. Additionally, due to the large scale of the environment, temporal condition expressions are crucial for agents to navigate effectively, as they are useful for describing long-distance travel.

\begin{table*}[t]
\begin{center}
\resizebox{1.85\columnwidth}{!}{
 \begin{tabular}{|l|c|c|c|c|c|c|c|c||c|c|c|c|c|c|c|}
  \hline
  \multicolumn{2}{|l|}{\multirow{4}{*}{Model}} & \multicolumn{7}{c||}{Val Seen} & \multicolumn{7}{c|}{Val Unseen}\\
  \cline{3-16} 
  \multicolumn{2}{|l|}{}&\multicolumn{5}{c|}{Navigation} & \multicolumn{2}{c||}{Assembly}&\multicolumn{5}{c|}{Navigation} & \multicolumn{2}{c|}{Assembly}\\
 \cline{3-16} 
 \multicolumn{2}{|l|}{} &\multirow{2}{*}{nDTW} & \multicolumn{4}{c|}{CTC}&

 \multirow{2}{*}{rPOD} &  \multirow{2}{*}{PTC}&\multirow{2}{*}{nDTW} & \multicolumn{4}{c|}{CTC}&

 \multirow{2}{*}{rPOD} &  \multirow{2}{*}{PTC}\\
 \cline{4-7} 
 \cline{11-14}
  \multicolumn{2}{|l|}{} &  & k=0 & k=3 & k=5 & k=7  &  & &  & k=0 & k=3 & k=5 & k=7  &  & \\
 \hline

    \multirow{3}{*}{V/L}  & T1& 0.222 & 0.000 & 0.138 & 0.194 & 0.260 &0.088  & 0.070
    &0.186 & 0.000 & 0.080 &0.139 & 0.192 & 0.054 &0.044\\
    \cline{2-16} 
    &T2 &0.049 &0.000  &0.057 & 0.103 &0.140 &0.027  &0.017
    &0.033 &0.000 & 0.044 & 0.078 &  0.113  &0.019 &0.011\\
    \cline{2-16}  
    &total &0.135 & 0.000 & 0.098 &0.149  & 0.200 & 0.058 &0.044
    & 0.109& 0.000& 0.062 & 0.108 & 0.153&  0.036  &0.028\\
    \hline

\end{tabular}
}
\end{center}
\vspace{-10pt}
\caption{Performance of Vision-and-Language (V/L) baseline for turns T1 and T2, plus overall scores on the Val-Seen/Unseen splits. \label{tbl:main_result_app}
} 
\vspace{-5pt}
\end{table*}

\begin{table}[t]
\begin{center}
\resizebox{0.95\columnwidth}{!}{
 \begin{tabular}{|l|c|c|c|c|c|c|c|}
  \hline

  \multirow{2}{*}{Model} &Navigation &
  \multicolumn{2}{c|}{Assembly}\\
  \cline{2-4} 
  &CTC (k=3)& rPOD& PTC\\
 \hline

    Vision-and-Language   &1.000  & 0.539 & 0.382\\
    \hline
\end{tabular}
}
\end{center}
\caption{Scores in the assembly phase calculated under the assumption of the perfect performance in the navigation phase on Val-Seen split.}
\label{tbl:Assebly_result}

\end{table}

\begin{figure*}[t]
    \centering
    \includegraphics[width=1.9\columnwidth]{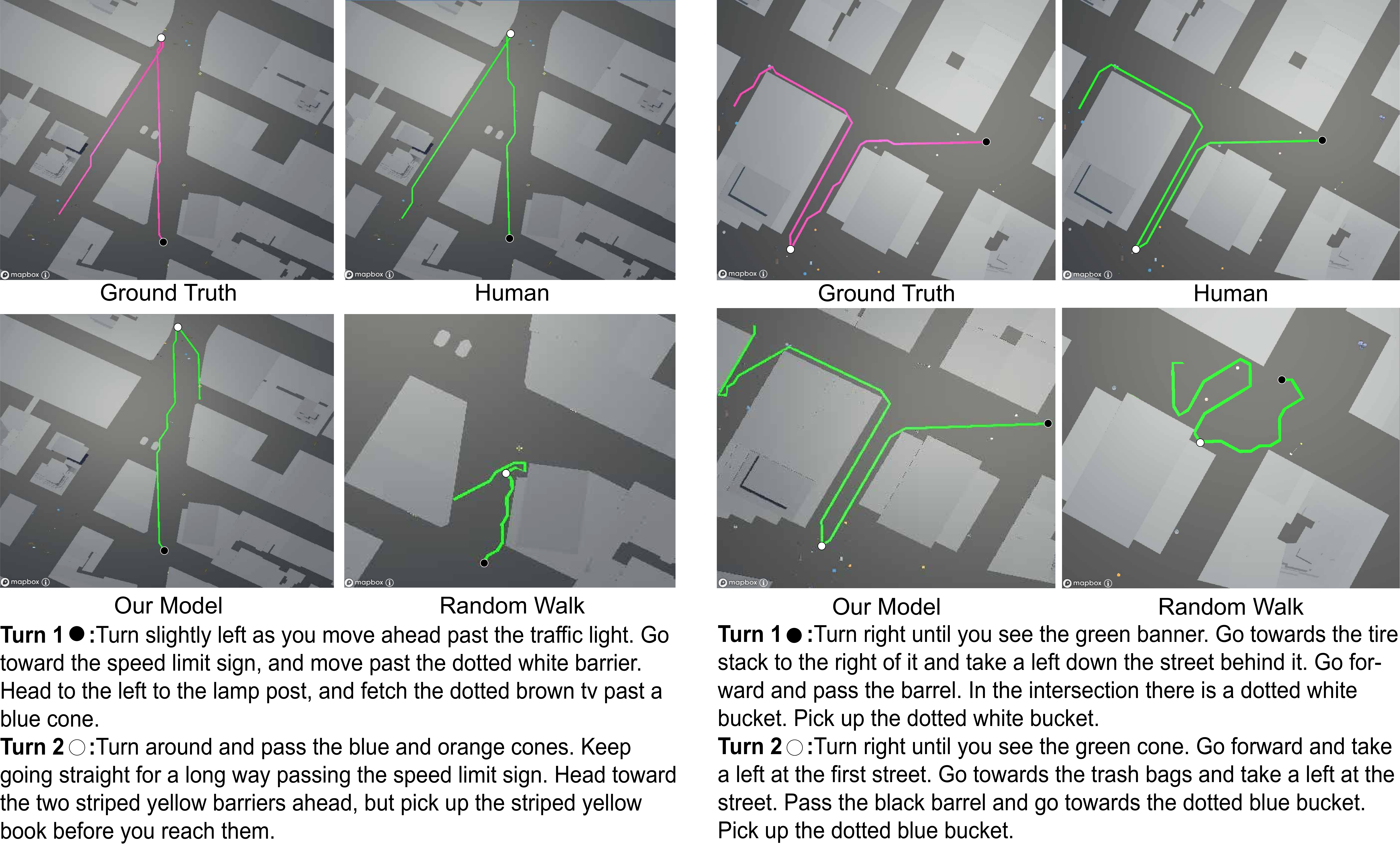}
    \caption{Navigation paths of ground truth, human evaluation, random walk, and our model. Pink is the GT path and the other paths are shown in green (turn 1 starts from the black dot and goes to the white dot. Turn 2 starts from white dot and goes to the end of the path).}
    
    \label{fig:vis2}
\end{figure*}

\begin{figure*}[t]
    \centering
    \includegraphics[width=1.9\columnwidth]{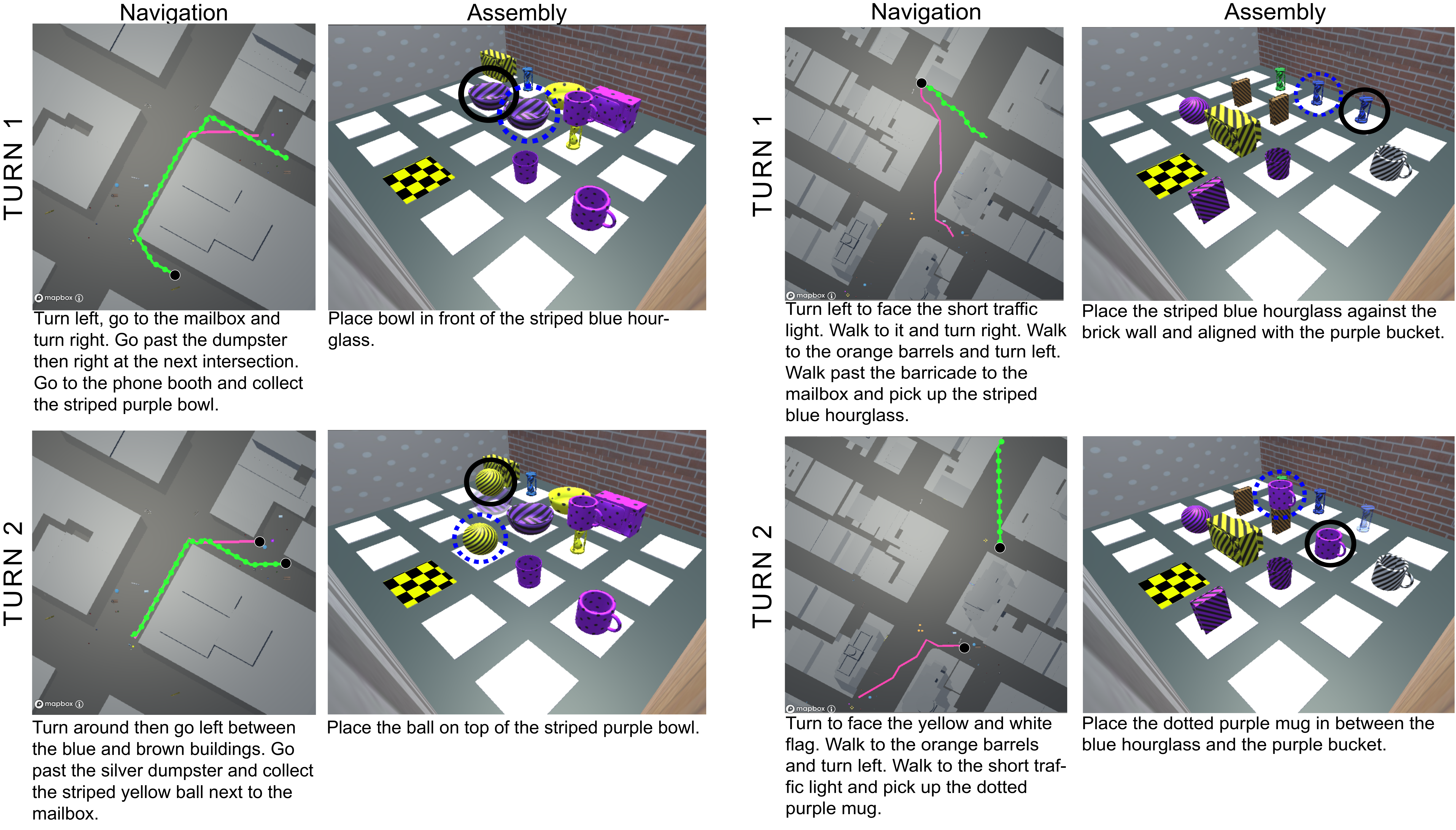}
    \caption{Visual demonstrations by our model in navigation and assembly phases. GT navigation paths are solid pink lines and model's paths are dotted green lines (start = black dot). GT assembly target location is solid black circle and model's target object placement is dashed blue circle (start = checkered yellow tile, agent facing brick wall).}
    
    \label{fig:vis2_assm}
    
\end{figure*}

\section{Model\label{app:models}}
\paragraph{Cross Attention.}
We employ the bidirectional attention mechanism \cite{seo2016bidirectional} to align the visual feature \(V\) and instruction feature \(L\).
We calculate the similarity matrix, \(S \in \mathbb{R}^{w' \times l} \) between visual and instruction.
\begin{equation}
    S_{ij} = W_s^{\top} (V_i \odot L_j)
\end{equation}
where \(W_s \in \mathbb{R}^{d \times 1}\) is the trainable parameter, and \(\odot \) is element-wise product. From the similarity matrix, the new fused instruction feature is:
\begin{align}
    \bar{V} &= \textrm{softmax}(S^{\top})V \\
    \hat{L} &= W_L^{\top}[L;\bar{V};L \odot \bar{V}]
\end{align}
Similarly, the new fused visual feature is:
\begin{align}
   \bar{L} &= \textrm{softmax}(S)L \\
    \hat{V} &= W_V^{\top}[V;\bar{L};V \odot \bar{L}]
\end{align}
where \(W_L\) and \(W_V\) are trainable parameters. 

\paragraph{General Attention.}
We employ a basic attention mechanism for aligning action feature, \(h\), and each of visual and instruction features.
\begin{align}
    A_{i} &= \hat{V}_i^{\top}h\\
    \alpha &= \textrm{softmax}(A) \\
    v &= \alpha^{\top}\hat{V} 
\end{align}

\section{Experiments}
\subsection{Simulator Setup} 
Our task is quite challenging. In many cases, agents may not even be able to pick up an object in the navigation phase (agents would have to be in a position close enough to the object and of the correct rotation to pick the object. These factors along with the size of the environment, make this difficult). To decrease the difficulty of the task, in the event agents do not successfully pick up an object, we allow them to continue to the assembly phase with whatever object is the closest to their final location. Likewise in the assembly phase, if the time step limit is reached before the agent places the object down, the object will be placed in front of them (in the event ``in front of them'' is out of bounds, it is placed at their feet). Note that either of these actions will result in PTC and rPOD to be 0.

\subsection{Training Details\label{app:train_detail}}
We use PyTorch \cite{paszke2017automatic} to build our model. We take the average of the losses from navigation and assembly phase modules to calculate the final loss. We use 128 as a hidden size of linear layers and LSTM. For word and action embedding sizes, we use 300 and 64, respectively. The visual feature map size is \(7\times7\) with 2048 channel size. For dropout p value, 0.3 is used. We use Adam \cite{kingma2014adam} as the
optimizer and set the learning rate to 0.001. The number of trainable parameters of our Vision-and-Language model is 1.83M (Language-only: 1.11M, Vision-only: 0.73M). We use NVIDIA RTX 2080 Ti and TITAN Xp for training and evaluation, respectively. 

\section{Results and Analysis\label{app:results}}
As shown in Table \ref{tbl:main_result_app}, almost all scores from turn 1 are improved compared to turn 2. Scoring in rPOD and PTC metrics in the assembly phase is largely dependent on the score of CTC-k in the navigation phase. Comparing the rPOD and PTC scores of Vision-and-Language model on the val-seen split (Table~\ref{tbl:main_result_app}) and the ones from Table~\ref{tbl:Assebly_result}, if the CTC-k is decreased by 1/10 (1.0 to 0.098), the PTC is also decreased around 1/10 (0.382 to 0.044). This demonstrates our \dataname{} task involves interweaving and is challenging to complete.

\section{Output Examples\label{app:output_examples}}
In the left path set of Figure \ref{fig:vis2}, our model follows the instructions well in the beginning. However, the model goes a little bit further and fails to find the target object (dotted brown tv). In the second turn, the model turns around, but does not do it fully, so heads a different direction failing to reach the goal position.

For the example on the right, the model performs very well in the first turn, but in the second turn fails to find the target object although reaches very close to it and then backtracks out of the alley.
Also, as shown in the figure, the human performs the navigation almost perfectly, indicating there is significant room for improvement by future work, and random-walk shows quite poor performance, implying that our \dataname{} task cannot be completed by random chance.
 
Figure \ref{fig:vis2_assm} compares the model against the GT in both turns and phases. On the left set, the model almost reaches the target object, but it cannot find the target object (striped purple bowl) and goes a little further past it. In the corresponding assembly phase, the model places the collected object (assuming it picked up the correct object in the previous navigation phase) 1 space to the right of the target location. In the next navigation turn, due to the error in the previous turn, the model path starts a bit further away from the GT, however, it starts to realign itself towards the end around the corner. The model is able to locate the target object and stop to pick it up. In the next assembly phase, the model fails to place the collected object at the right location. On the right set, the model shows worse performance. It misses all of the turning needed to reach the target. In the assembly phase, the model misses the target location by 1 space, likely due to misunderstanding the complex spatial relationship in the instructions. In the next navigation phase, the model starts in the wrong place, so ends up arriving at a totally different place from the target position. In the next assembly phase, the performance of the previous turn affected the object configuration, so the model cannot find the place ``between the blue hourglass and the purple bucket".

\end{document}